\documentclass{article}

\usepackage{PRIMEarxiv}
\usepackage[utf8]{inputenc} % allow utf-8 input
\usepackage[T1]{fontenc}    % use 8-bit T1 fonts
\usepackage{hyperref}       % hyperlinks
\usepackage{url}            % simple URL typesetting
\usepackage{booktabs}       % professional-quality tables
\usepackage{amsfonts}       % blackboard math symbols
\usepackage{nicefrac}       % compact symbols for 1/2, etc.
\usepackage{microtype}      % microtypography
\usepackage{lipsum}
\usepackage{fancyhdr}       % header
\usepackage{graphicx}       % graphics
\usepackage{subfig}
\usepackage{lscape}
\usepackage{multirow}
\usepackage{bm}
\usepackage{arydshln}
\usepackage{amsmath}
\graphicspath{{media/}}     % organize your images and other figures under media/ folder

\title{Spatiotemporal forecasting of vertical 
track alignment with exogenous factors
\thanks{\textit{\underline{Citation}}: 
\textbf{Katsuya Kosukegawa, Yasukuni Mori, Hiroki Suyari, Kazuhiko Kawamoto, Spatiotemporal forecasting of vertical track alignment with exogenous factors, Scientific Reports 13, 2354, 2023.
DOI:10.1038/s41598-023-29303-7}} 
}

\author{
  Katsuya Kosukegawa, Yasukuni Mori, Hiroki Suyari, Kazuhiko Kawamoto \\
  Chiba University, Japan \\
  \texttt{katsuya\_kosukegawa@chiba-u.jp},
  \texttt{\{yasukuni,suyari,kawa\}@faculty.chiba-u.jp} 
}

\begin{document}
\maketitle

\begin{abstract}
To ensure the safety of railroad operations, it is important to monitor and forecast track geometry irregularities.
A higher safety requires forecasting with higher spatiotemporal frequencies, which in turn requires capturing spatial correlations.
Additionally, track geometry irregularities are influenced by multiple exogenous factors.
In this study, a method is proposed to forecast one type of track geometry irregularity, vertical alignment, by incorporating spatial and exogenous factor calculations.
The proposed method embeds exogenous factors and captures spatiotemporal correlations using a convolutional long short-term memory.
The proposed method is also experimentally compared with other methods in terms of the forecasting performance.
Additionally, an ablation study on exogenous factors is conducted to examine their individual contributions to the forecasting performance.
The results reveal that spatial calculations and maintenance record data improve the forecasting of vertical alignment.
\end{abstract}

% keywords can be removed
\section*{Introduction}
Tokaido Shinkansen is a high-speed train linking Japan's major cities: Tokyo, Nagoya, Kyoto, and Osaka.
Serious train accidents caused by track geometry irregularities have not occurred because of the high spatiotemporal frequencies of track maintenance by railroad operators.
Track maintenance activities can be categorized into two main groups: preventive and corrective maintenance. 
Preventive maintenance predicts failures before they occur, whereas corrective maintenance rectifies existing defects.
Currently, Tokaido Shinkansen operates under corrective maintenance.
A railroad operator performs track maintenance when the track conditions exceed the designed maintenance level.
The designed maintenance levels include precautionary and dangerous levels.
After identifying that the track condition exceeds the precautionary level, railroad operators plan track maintenance to be conducted within the next few days.
In contrast, if the track condition exceeds the dangerous level, railroad operators immediately perform emergency track maintenance.
Emergency track maintenance requires additional personnel and resources, resulting in additional costs.
Preventive maintenance can reduce these costs.
Suppose that the Tokaido Shinkansen operates on preventive maintenance, in this case, railroad operators can forecast whether the track condition will exceed the maintenance level.
Subsequently, railroad operators can perform track maintenance at points where track conditions are likely to exceed maintenance levels.
Consequently, the number of points where the track conditions exceed the preventive or dangerous levels is reduced.
Preventive maintenance guarantees railroad security and reduces costs by smoothing the amount of maintenance.

In the Tokaido Shinkansen, a track-inspection train, called ``Doctor Yellow’’, inspects track geometry irregularities,
which are track deformations.

\begin{figure}[tb]
    \centering
    \includegraphics[width=0.8\linewidth]{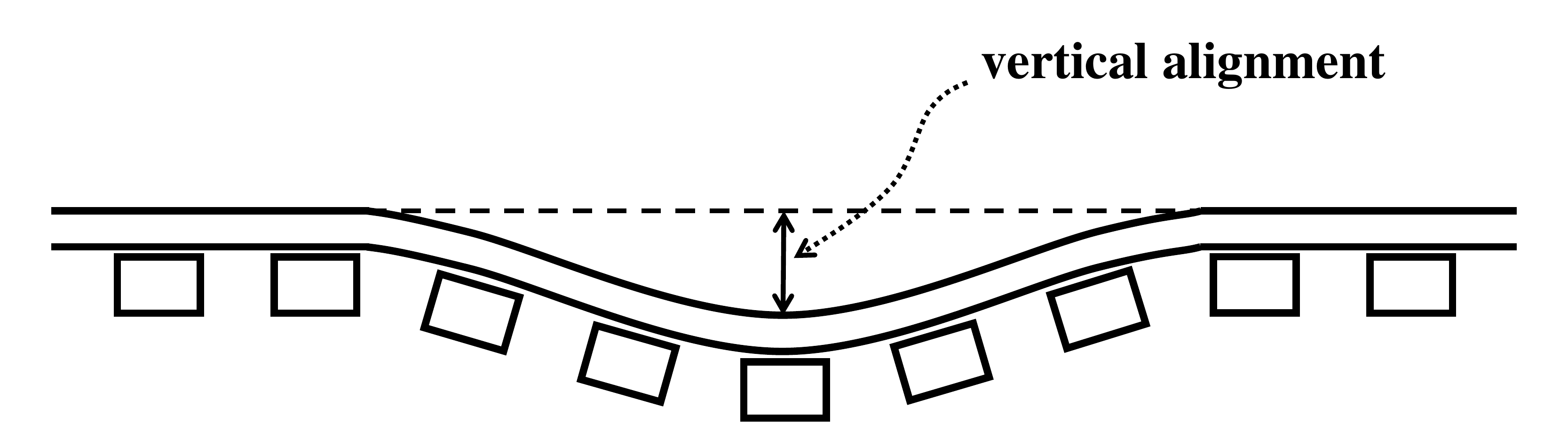}
    \caption{Vertical alignment.}
    \label{fig:vertical_alignment}
\end{figure}

Doctor Yellow inspects seven types of track geometry irregularities: two vertical alignments for left and right tracks, 
two lateral alignments for left and right tracks, 
gauge deviation, cross level, and twist.
Gauge is the distance between the right and left tracks.
%Doctor Yellow inspects the vertical and lateral alignments of each left and right track. 
Vertical alignment is critical among track geometry irregularities because it deteriorates ride quality and can cause train accidents.
Furthermore, in Japan, vertical alignment is used to decide whether maintenance operations need to be performed.
Figure \ref{fig:vertical_alignment} illustrates vertical alignment.
Additionally, Doctor Yellow measures the running speed 
and vehicle vibrations in the vertical and lateral directions.
Table \ref{tab:track_irregularity} describes the seven track geometry irregularities
and the three measurements. 
Henceforth, for simplicity, the seven irregularities
and the three measurements are collectively referred to as track geometry irregularities.

\begin{table}[tb]
\centering
\begin{tabular}{|l|l|}
\hline
Track geometry irregularity & Description \\
\hline\hline
\begin{tabular}{l}
Vertical alignments\\ for the left and right tracks
\end{tabular}
& Vertical displacement in the longitudinal direction of the 
left or right rail \\
\hline
\begin{tabular}{l}
Lateral alignments\\ for the left and right tracks
\end{tabular}
& Lateral displacement in the longitudinal direction of the 
left or right rail \\
\hline
Gauge deviation & Error between the design and actual measured value of the gauge \\
\hline
Cross-level & Deviation between the top surfaces of the left and right tracks at a given location\\
\hline 
Twist & Difference between two cross-level measurements taken a specific distance apart\\
\hline
Vertical vehicle vibration & Vertical vibration of the track-inspection train (Doctor Yellow)\\
\hline
Lateral vehicle vibration & Lateral vibration of the track-inspection train (Doctor Yellow)\\
\hline
Vehicle running speed & Running speed of the track-inspection train (Doctor Yellow)\\
\hline
\end{tabular}
\caption{\label{tab:track_irregularity}Track geometry irregularities.}
\end{table}

In some countries, track maintenance plans are determined
based on the standard deviation of the vertical alignment for each 100[m] or 200[m] section~\cite{Soleimanmeigouni:Track:2018,setiawan2016track}.
In contrast, Tokaido Shinkansen railroad operators manage tracks with high spatial and temporal frequencies to ensure safety.
The railroad operators inspect every 1[m] section approximately every 10 days with Doctor Yellow.
The inspection is used for planning the track maintenance schedule.
Track geometry irregularities have spatial correlations owing to the chord offset method used by Doctor Yellow.
The chord offset method measures the vertical and lateral alignments relative to the neighboring points and not the absolute values.
Thus, spatial correlations must be captured for forecasting the vertical alignment.
Additionally, the degradation of track geometry irregularities is affected by external factors such as rainfall, ballast deterioration, passing weight, and maintenance.
In particular, some maintenance restores the vertical alignment, resulting in jump changes in the vertical alignment.

This study proposes a convolutional long short-term memory (ConvLSTM)~\cite{shi2015convolutional} to forecast vertical alignment at
high spatial and temporal frequencies.
The proposed ConvLSTM design considers the spatial correlation and the effect of exogenous factors on vertical alignment.
The main contributions of the proposed method are summarized as follows.
\begin{itemize}
    \item The proposed method can forecast vertical alignment at high spatiotemporal frequencies using the exogenous data. 
    The forecasting of vertical alignment is of practical significance because it is used to decide whether to perform the maintenance operations of railroads in Japan.
    \item The proposed method extends the basic ConvLSTM to capture the spatial correlations and the effect of exogenous factors.
\end{itemize}

\section*{Related work}
Table \ref{tab:prior_research} summarizes prior research that predicted the vertical alignment.
Each column in the table represents the following item.
The Spatial and Temporal Frequency column shows the spatial and temporal frequency of the predicted features.
The Spatial Calculation column
indicates whether the model 
uses spatial calculation with neighboring spatial points or sections
The Exogenous Factor column indicates whether the model uses exogenous factors.
The Model column shows the machine learning model used in the study.
The Predicted Feature column shows the features predicted in 
the study.
The Forecast column indicates whether the study
forecasts the features.

    Sresakoolchai and Kaewunruen\cite{sresakoolchai2022railway} detect track component defects using track geometry. They use
    supervised learning techniques such as a deep neural network, a convolutional neural network (CNN),
    multiple regression, the support vector machine, gradient boosting, a decision tree, and random forests.
    Moreover, they use unsupervised learning techniques such as k-means clustering and association rules
    to explore and analyze the insights of track component defects.

Sadeghi et al.~\cite{sadeghi2010development}  predicts a track degradation coefficient $\textrm{TGI}_2/\textrm{TGI}_1$, the rate of the future track geometry index $\textrm{TGI}_2$ and the present track geometry index $\textrm{TGI}_1$.
\textrm{TGI} is the index based the standard deviation of track geometry irregularities.
Xu et al.~\cite{xu2012novel} approximated the nonlinear deterioration process of a surface (similar to vertical alignment) with many short linear processes.
The time measurement intervals were not constant; however, the mean was approximately 18 days.
The spatial interval of forecasting was 0.5[m].
Moreover, their method is linear and does not include spatial computations, in contrast to our method, which is nonlinear and includes spatial computations.
Soleimanmeigouni et al.~\cite{soleimanmeigouni2018modelling} proposed a track geometry irregularity degradation model that considers recovering the track geometry irregularities and the corresponding changes in the degradation rate.
The degradation rate changes with each maintenance action.
Recovery after maintenance is computed by a linear model that takes the track geometry irregularities before maintenance and the types of maintenance as the input. 
They also argued that the initial values of the track geometry irregularities and the corresponding degradation rate have spatial correlations, which were then computed using the ARMA model. 
Soleimanmeigouni et al. selected a model with reversibility (the ARMA model) because spatial relationships have a two-directional dependence.
The ARMA model computes only the spatial correlations in the initial state.
In contrast, the proposed model computes spatial correlations for the degradation process.
Maintenance and other exogenous factors have a spatial effect on track geometry irregularities, particularly because of the chord offset method.
The spatial measurement interval of the aforementioned ARMA model is 200[m]
and the corresponding spatial frequency is lower than that of our method.
Notably, the ARMA model allowed for any temporal frequency as time was included as an independent variable.
Chang et al. use a multi-stage linear model to forecast $\textrm{SD}_{LL}$, the standard deviation of the longitudinal level (same as the vertical alignment) between two maintenance.
They use the total passing tonnage from last maintenance as explanatory variable.
Soleimanmeigouni et al.~\cite{soleimanmeigouni2020prediction} used logistic regression to predict the occurrence of longitudinal isolated defects, which are 
short irregularities in the track geometry. We forecast vertical alignment with high spatial frequency to detect local defects. 
In Tokaido Shinkansen, the railroad operators focus on preventive monitoring and repair of local defects for safety.
Goodarzi et al.~\cite{GOODARZI2022128166} also used logistic regression and gradient boost machine (GBM) to predict yellow-tag defects at the next inspection (YDNI). 
A yellow-tag defect indicates that the track geometry will soon become a red-tag defect.
The red-tag defect indicates that the track geometry violates the Federal Railroad Administration (FRA) track safety standards and must be corrected as soon as possible\cite{cardenas2017ensemble}.
They use the train load in million gross tonnes (MGT), track class, and ballast fouling index (BFI) as input data.
The result shows that the BFI improves 
the forecasting performance.
This study uses the ballast age instead of the ballast deterioration.
Movaghar and Mohammadzadeh~\cite{movaghar2022bayesian} computed a Bayesian linear regression model with exogenous data as the independent variable and $\textrm{SD}_{\textrm{LL}}$, the standard deviation of the longitudinal level (same as the vertical alignment), as the dependent variable.
Their method was applied to each track section with track section lengths of 13, 23, 24, and 18 [km].
The spatial frequency of our method is higher than that of their method.
Moreover, although the temporal frequency of their forecasting method is one year, it can forecast at any temporal frequency because the method includes time as an independent variable.

Guler~\cite{guler2014prediction} uses artificial neural networks (ANNs) to predict the degradation rate of the track geometry irregularities between two maintenance works for each analytical segment (AS). 
The average length of AS is 220 m.
They input exogenous data such as traffic loads and speed
into the ANN.
This study also uses the passing tonnage since the last inspection as exogenous data.
Chen et al.~\cite{chen2021learn} used long short-term memory (LSTM)~\cite{LSTM}, gated recurrent units (GRU)~\cite{GRU}, CNNs, and their ensembles to forecast vertical track irregularities (same as vertical alignment) in time and spatial series.
Their temporal measurement interval was one month, and the time frequency of the forecast was one year.
Their spatial measurement interval was 0.5[m].
Their method does not include spatial computations in time-series forecasting, in contrast to our method, which includes spatial computations.
Additionally, the temporal frequency of our method is higher than theirs.
Khajehei et al.~\cite{khajehei2022prediction} used ANN to predict the degradation rate of $\textrm{SD}_{\textrm{LL}}$.
They fed a variety of exogenous data into the model.
They used the ballast age, the maintenance history, the level of degradation after tamping or renewal,  the average annual frequency of trains, etc.
They used the Garson method to investigate the relative importance of the variables affecting the rate of geometry degradation.
They found that the maintenance history made the most significant contribution.
This study also uses maintenance records as exogenous data.
Sresakoolchai and Kaewunruen\cite{sresakoolchai2022track} used recurrent neural networks (RNN), LSTM, GRU, and an attention mechanism to forecast seven track geometry irregularities. 
They proposed a model that computes spatial relationships.
However, the model only computes unidirectional spatial neighborhoods, even though spatial relations
are bidirectional.
This study uses CNN within ConvLSTM to compute bidirectional spatial relations.

\begin{landscape}
\begin{table}[t]
\centering
\begin{tabular}{|l|c|c|c|c|c|c|c|}
\hline
\multirow{2}{*}{Reference} & \multicolumn{2}{|c|}{Frequency} & \multirow{2}{*}{\shortstack{Spatial\\ calculation}} & \multirow{2}{*}{\shortstack{Exogenous\\ factor}} & \multirow{2}{*}{Model} & \multirow{2}{*}{Predicted feature} & \multirow{2}{*}{Forecast} \\
\cline{2-3}
 & Spatial & Temporal &  &  & & & \\
\hline
Sadeghi et al. \cite{sadeghi2010development} & 0.6 km (mean) & any &  & \checkmark & exponetial regression & track degradation coefficient & \checkmark \\
Xu et al.\cite{xu2012novel} & 0.5 m & 18 days (mean) &  & & linear model & gage, right Surface & \checkmark \\
Soleimanmeigouni et al.\cite{soleimanmeigouni2018modelling} & 200 m & any & \checkmark & \checkmark & piecewise line model, ARMA & $\textrm{SD}_{\textrm{LL}}$ level & \checkmark \\
Chang et al. \cite{chang2010multi} & 200m & any & & \checkmark & multi-stage linear model & $\textrm{SD}_{\textrm{LL}}$ & \checkmark \\
Soleimanmeigouni et al.\cite{soleimanmeigouni2020prediction} & 100\textasciitilde 300 m & - &  &  & logistic regresssion & isolated longitudinal level defect &  \\
Goodarzi et al.\cite{GOODARZI2022128166} & 200 ft & next inspection &  & \checkmark & logistic regression, GBM & YDNI & \checkmark \\
Movaghar et al.\cite{movaghar2022bayesian} & 13 km\textasciitilde & any & &  & Bayesian linear regression & $\Delta\textrm{SD}_{\textrm{LL}}$ & \checkmark \\
Guler\cite{guler2014prediction} & 220 m (mean) & - &  & \checkmark & ANN & deterioration rate & \\
Chen et al.\cite{chen2021learn} & 0.25 m & 1 month & & ? & ARIMAX, LSTM, GRU, CNN & vertical track irregularity & \checkmark \\
Khajehei et al.\cite{khajehei2022prediction} & 100\textasciitilde300 m & - &  & \checkmark & ANN & degradation rate of $\textrm{SD}_{\textrm{LL}}$& \\
Sresakoolchai et al.\cite{sresakoolchai2022track} & 1 ft & 1 year & \checkmark & \checkmark & RNN, LSTM, GRU, Attention & 7 types of track geometry  & \checkmark \\
Ours & 1 m & 10 days & \checkmark & \checkmark & ConvLSTM & vertical alignment & \checkmark \\
\hline
\end{tabular}
\caption{\label{tab:prior_research}Vertical alignment prediction models.}
\end{table}
\end{landscape}

\section*{Dataset}
Tables \ref{tab:track_irregularity} and \ref{tab:data_type} summarize the track geometry irregularity data and exogenous data, respectively.
These data are provided by the Central Japan Railway Company.
This study aims to identify and verify the exogenous factors that are significant for forecasting vertical alignment.

\subsection*{Track geometry irregularity}
As explained in the Introduction, track geometry irregularity refers to the track deformation.
As listed in Table \ref{tab:track_irregularity}, the track geometry irregularity has seven types of deformations. 
Moreover, Doctor Yellow measures the running speed
as well as the vertical and lateral vehicle vibrations.
Henceforth, these 10 total data categories are collectively referred to as track geometry irregularity data.

Vertical and lateral alignments for each left and right track
are measured using
the 10[m] chord offset method (also called the 10[m] versine method)~\cite{naganuma2010inertial}.
This method measures the relative alignments of adjacent points that are 5[m] apart.
Relative alignments $v(l)$ and absolute alignments $u(l)$ at 
spatial position $l$ are
related by
\begin{equation}
    \label{eq:chord_offset_method}
    v(l)=u(l)-\frac{u(l-5)+u(l+5)}{2}+\varepsilon,
\end{equation}
where, $\varepsilon$ denotes the measurement error.
The relative alignments $u(l)$ are used as input to the proposed model,
because the absolute alignments $u(l)$ cannot be measured.
Equation \ref{eq:chord_offset_method} indicates that the vertical and lateral alignments are
spatially correlated, demonstrating the need for a forecasting method that considers spatial correlations.

\subsection*{Exogenous data}
    Table \ref{tab:data_type} lists the exogenous data.
    In addition to Table \ref{tab:data_type}, there are more exogenous factors that affect track alignment.
    For example,
    the ballast and the soil conditions are thought to strongly
    affect vertical alignment.
    However, such conditions
    are difficult to directly measure using a high-speed inspection train.
    Furthermore, turnouts may affect track alignment but are not
    considered because the available data include limited turnouts.
    In this study, only the observable data is used to forecast
    vertical alignment.
    For example, the ballast age is used instead of the ballast condition. 
    The exogenous data are discussed in detail next.
    
    \begin{table}[t]
    \centering
    \begin{tabular}{|l|l||c|c|l|}
    \hline
    \multicolumn{2}{|l||}{Data} & Spatial & Temporal & Data type \\
    \hline\hline
    \multirow{6}{*}{Exogenous} & maintenance record & \checkmark & \checkmark & binary \\
    \cline{2-5}
    & under-structure & \checkmark & & categorical\\
    \cline{2-5}
    & rail joint & \checkmark & & binary \\
    \cline{2-5}
    & ballast age & \checkmark & \checkmark & real number \\
    \cline{2-5}
    & tonnage & \checkmark & \checkmark & real number\\
    \cline{2-5}
    & rainfall & \checkmark & \checkmark & real number\\
    \hline
    \end{tabular}
    \caption{\label{tab:data_type}Exogenous data.}
    \end{table}

\noindent{\bf Maintenance record}: 
These data represent the records
of track maintenance performed by  railroad operators.
Table \ref{tab:maintenance_list} lists
the maintenance operations used by railroad operators for correcting
vertical alignment.
In the list, the categories of sleeper maintenance and others comprise more detailed operations (See Supplementary 
Table S1 online).
For example, sleeper maintenance includes 
sleeper replacement, loose sleeper repair, and so on.
The aforementioned detailed operations 
are merged because of limited data.
The data for the maintenance operations
in Table \ref{tab:maintenance_list} is used to forecast vertical alignment.
Subsequently, each selected maintenance is represented as a binary value that indicates whether the maintenance is scheduled before the next track geometry irregularity inspection date.
Some maintenance operations directly restore vertical alignment, whereas others indirectly affect it.

\begin{table}[t]
    \centering
    \begin{tabular}{|ccc|}
        \hline
        \multicolumn{3}{|c|}{Maintenance}\\
        \hline \hline
         Uneven fixing & Tamping by multiple tie tamper & Manual tamping\\
         Ballast replacement  & Right rail replacement  & Left rail replacement\\
         Sleeper maintenance & Remediation of mud-pumping & Others\\
         \hline
    \end{tabular}
    \caption{\label{tab:maintenance_list}Maintenance operations for vertical alignment correction}
\end{table}

\noindent{\bf Under-structure}:
These data indicate five types of the structure and topography under the track: bridge, tunnel, overpass, embankment, and excavation.
Normally, track geometry irregularities tend to degrade near the boundaries of the structures.
The corresponding data are represented as a categorical variable that indicates which of the five aforementioned structural types categorizes every spatial point.

\noindent{\bf Rail joint}:
These data indicate the positions and types of rail joints.
The joint types are fourfold: insulated, welded, and expansions 
for right and left rail.
Track geometry irregularities also tend to occur around the joints.
The corresponding data are represented as binary variables that indicate 
which of the three aforementioned joint types categorize the joint positions.

\noindent{\bf Ballast age}:
These data represent the elapsed time since the ballast (crushed stone under rails) was replaced or installed.
Ballast deterioration affects the rate of vertical alignment deterioration~\cite{Soleimanmeigouni:Track:2018}.
However, ballast deterioration is
difficult to measure using
the high-speed track-inspection train.
Instead, the ballast age is inputted into the 
proposed model, assuming that the ballast age
is a linear approximation of the ballast deterioration.
The bridge section of the Tokaido Shinkansen is not a ballast track.
Moreover, the vertical alignment is not degraded in the sections without a ballast track.
Therefore, assuming that the track at the bridge section is like a new ballast track, the ballast age of the bridge section is defined as zero years old.

\noindent{\bf Tonnage}:
These data indicate the total weight of vehicles that have passed through the inspected point since the last date of track geometry irregularity inspection.
The greater the passing tonnage, the more the track geometry irregularity degrades~\cite{guler2014prediction}.

\noindent{\bf Rainfall}:
These data indicate the precipitation since the date of the last track irregularity inspection.
Rain degrades the soil condition under the track
and indirectly affects track geometry irregularities~\cite{NARVELYNGBY:Railway:177}.
The precipitation is considered as an exogenous factor instead of the soil condition.
Specifically, the cumulative precipitation and the maximum unit time rainfall are calculated from the last inspection date.
Accumulated precipitation indicates long-term rainfall, such as that during rainy season. In contrast, the maximum unit time precipitation indicates short-term rainfall, such as a sudden downpour.
The unit times for precipitation are 10 min, 1 hour, and 1 day.
Therefore, the rainfall data types are fourfold: accumulated precipitation, maximum of 10 min rainfall, maximum of hourly rainfall, and maximum of daily rainfall.

\subsection*{Data collection}
Data are collected from a 15[km] section of the Tokaido Shinkansen between 27[km] and 42[km] of the outbound line
from April 2011 to June 2021.
Track geometry irregularities are measured at 25[cm] intervals and then aligned to 1[m] intervals.
Therefore, track geometry irregularities are measured at 15,000 points in the spatial dimension, where
the number of spatial measurement points are denoted by $L=15,000$.
The exogenous data are also aligned at 1[m] intervals to match the track geometry irregularity data.
    For the data, the odometry error is negligible due to high spatial resolution measurements and the odometry correction.

The track geometry irregularities are measured at unequal intervals approximately every 10 days.
The exogenous data are aligned with the measurement date of track geometry irregularities.
The dataset is split into training, validation, and test datasets according to the time dimension.
Data from 04/06/2011 to 04/26/2017, 05/08/2017 to 11/07/2018, and 11/19/2018 to 06/18/2021 are used as the training, validation, and test datasets, respectively,
where 
the number of sequences in the dataset of interest at that time are denoted as $T$.
Sliding-window processing is applied to each dataset and then each dataset is divided into sequences.
Next, the sequences are inputted into the model during training and inference.

\section*{Methods}

\subsection*{Formulation}
    Using track geometry irregularities and exogenous data,
    a spatiotemporal model is developed for forecasting the vertical alignments of the Tokaido Shinkansen rail.
    The track geometry irregularity data at time $t$ are denoted as 
    \begin{eqnarray}
        \label{eq:formulate_x}
        {x}_t=\left\{x_{l,c_x}^{(t)} \mid l=1\ldots L,c_x=1\ldots C_x\right\} 
    \end{eqnarray}
    where $x_{l,c_x}^{(t)}$ is the track geometry irregularity measurement at spatial position $l$ for irregularity type $c_x$.
    The number of spatial measurements is $L=15,000$, and
    $C_x=10$ types of irregularities exist.
    The vertical alignments at time $t$ are denoted as 
    \begin{eqnarray}
        \label{eq:formulate_y}
        {y}_{t}=\left\{y_{l,c_y}^{(t)} \mid l=1\ldots L,c_y=1\ldots C_y\right\}
    \end{eqnarray}
    where $y^{(t)}_{l,c_y}$ denotes the vertical alignment at the 
    spatial position $l$ and $c_y$ is an index indicating whether
    the left or right track, that is, $C_y=2$.
    The objective is to build a 
    vertical alignment forecast model $f$ as
    \begin{equation}
        \hat{y}_{t+1}=f\left({x}_{t-\tau+1:t},{e};{\theta}\right)
    \end{equation}
    where ${x}_{t-\tau+1:t}=\left\{{x}_{t-\tau+1}\ldots {x}_{t}\right\}$ is the input sequence of the track geometry irregularities. 
    Moreover, ${e}$ denotes exogenous data,
    and $\theta$ are the parameters of model $f$.
    Exogenous data ${e}$ include several types of data, as summarized in Table \ref{tab:data_type}.
    To find an optimal
    parameter $\hat{\theta}$,
    the following minimization problem is formulated:
    \begin{equation}
        \hat{\theta}=\underset{\theta}{\textrm{argmin}}\left\|{y}_{t+1}-f\left({x}_{t-\tau+1:t},{e};{\theta}\right)\right\|^2.
        \label{eq:minimize}
    \end{equation}

\subsection*{Proposed model}
\begin{figure}[tb]
    \centering
    \includegraphics[width=\linewidth]{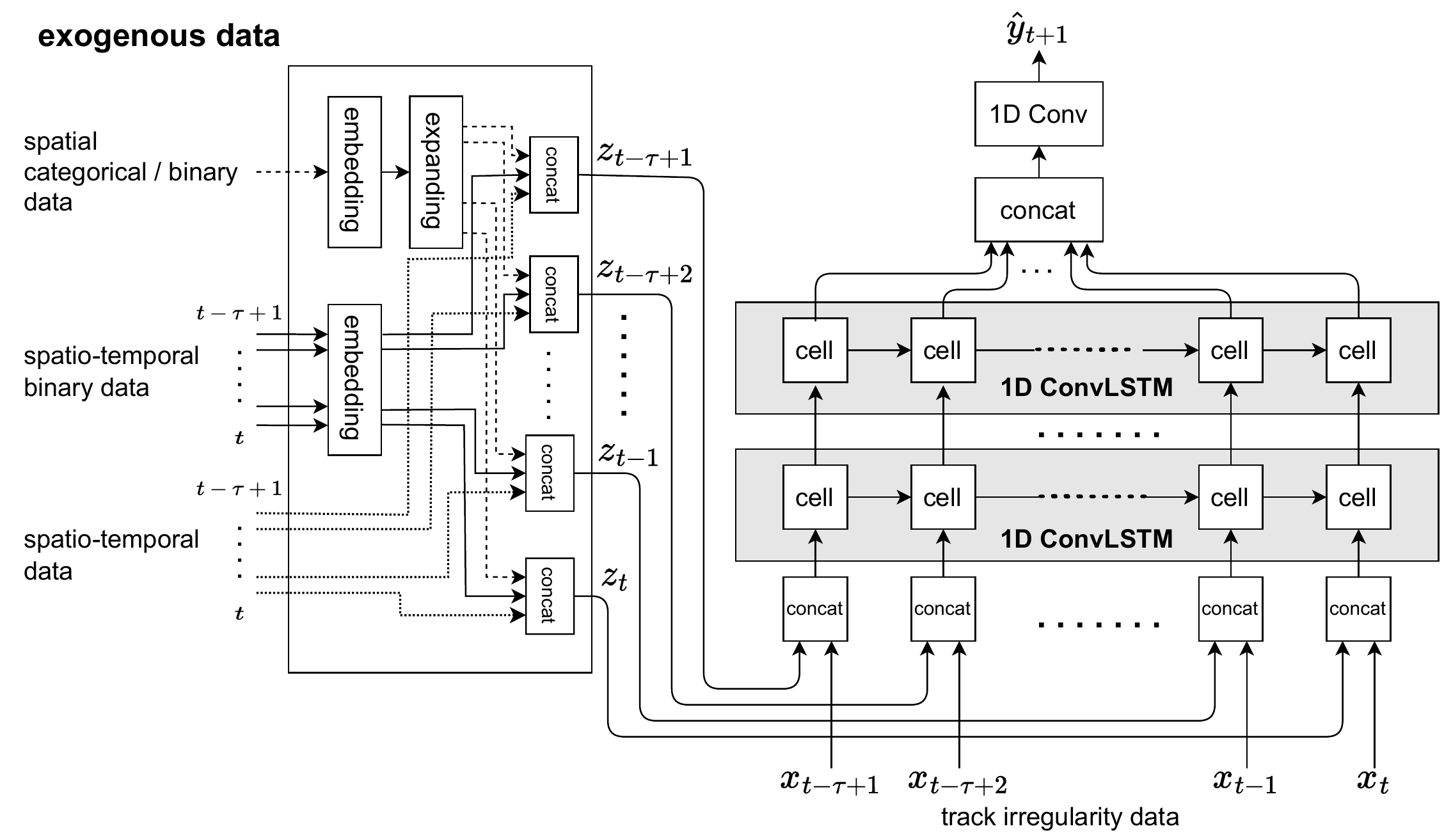}
    \caption{ConvLSTM with exogenous data, i.e., the proposed model that predicts vertical alignment. The left part is the exogenous embedding part. The right part is the ConvLSTM part.}
    \label{fig:proposed_model}
\end{figure}

A model $f$ is developed for forecasting vertical alignment based on
the ConvLSTM \cite{shi2015convolutional}.
Figure \ref{fig:proposed_model} shows the proposed forecasting model $f$
with one-dimensional (1D) ConvLSTM cells and an embedding part for exogenous data.
The exogenous embedding part computes 
embedding features ${\bm z}_{t-\tau+1},\ldots, {\bm z}_{t}$
from the exogenous data.
The 1D ConvLSTM cells are used to forecast the vertical alignment
from the track geometry irregularities and the
exogenous embedding features.

\begin{figure}[t]
    \centering
    \includegraphics[width=0.7\linewidth]{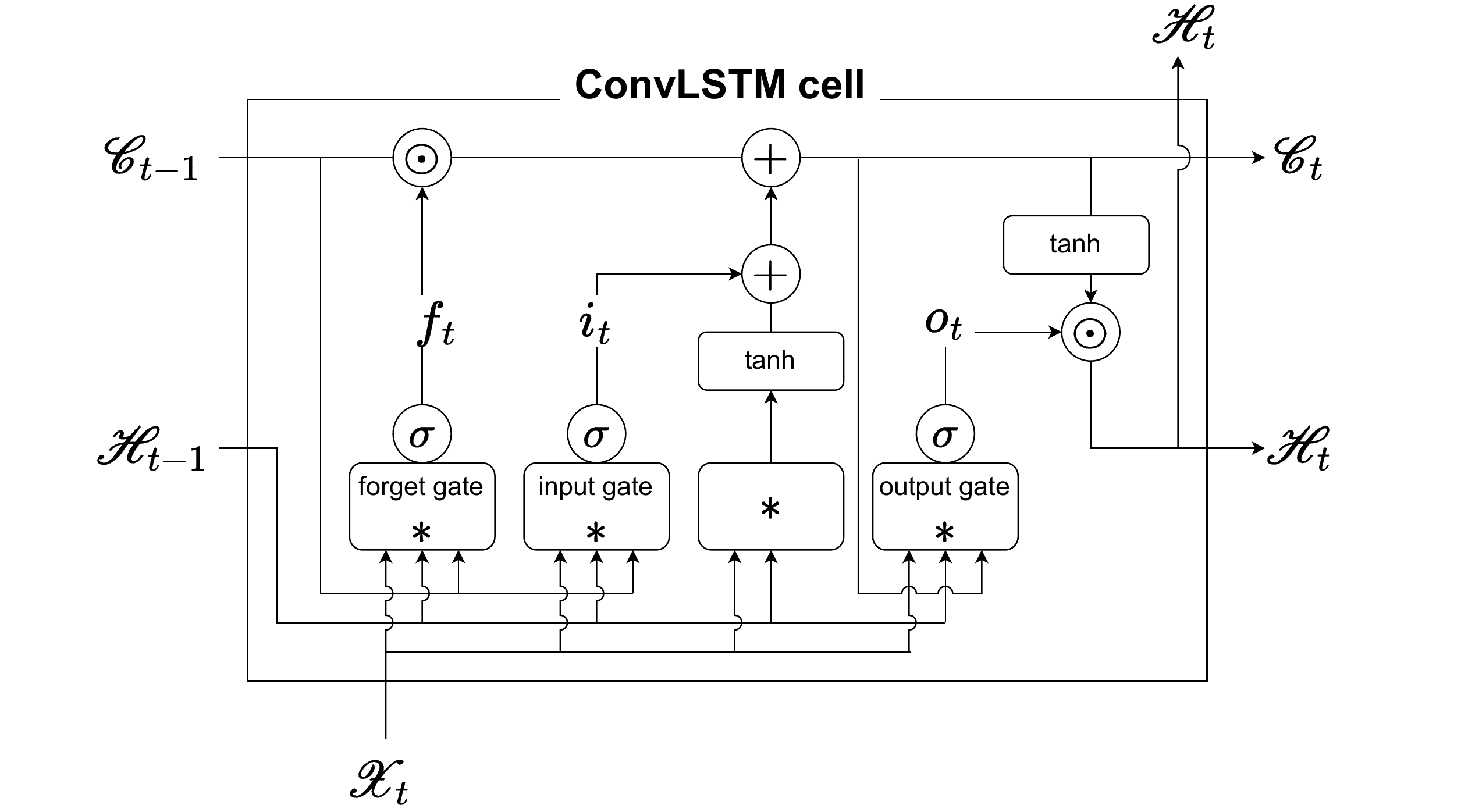}
    \caption{ConvLSTM cell.}
    \label{fig:convlstm_cell}
\end{figure}

\noindent{\bf Exogenous embedding}:
The exogenous embedding part in Figure \ref{fig:proposed_model}
consists of embedding layers, expanding layers, and concatenation
layers.
The embedding layers map the sparse representations
of the exogenous data to dense feature vectors.
The proposed model has an embedding layer for each data format in Table \ref{tab:data_type}.
Each embedding layer is a fully-connected network (FCN)
with the input size depending on the input data format,
whereas the output of FCN is a 
4D vector for all data formats.
The expanding and concatenate layers are used to
obtain the tensor representation of the exogenous features ${\bm z}_{t}$ from a set of the 4D embedding vectors
for all data formats.
Next, the way of embedding
for each data format in Table \ref{tab:data_type} are discussed.

For spatiotemporal binary data (maintenance record), a maintenance record at spatial position $l$ at time $t$
is represented as binary value ($0$ or $1$), where $1$ indicates that the maintenance 
operation is performed.
When feeding the record into FCN,
the binary representation of $0$ and $1$ is slightly modified as
one-hot vector $(0,1)$ and $(1,0)$, respectively.
The one-hot representation is necessary to avoid zero-input; 
training FCN becomes impossible because the output
to zero-input is zero.
For each maintenance record at spatial position $l$ at time $t$,
FCN outputs a 4D embedding vector.
By repeating the embedding computation over the spatiotemporal
and categorical domain, 
a set of the embedding vectors is obtained.
Finally,
the set of the embedding vectors is concatenated in the spatial, temporal, and categorical directions to obtain the tensor representation.

For spatial categorical data (under-structure),
five types of under-structure
are represented at spatial position $l$
as a 5D one-hot vector, e.g.,~$(0,1,0,0,0)$.
For each under-structure at spatial position $l$,
FCN for the corresponding data outputs a 4D embedding vector.
The embedding vector is expanded in the temporal direction
to obtain the tensor representation
with the same length along the temporal axis
as that of the tensor of the spatiotemporal categorical data.
Finally, the set of the expanded embedding vectors is 
concatenated in the spatial direction
to obtain the tensor representation.
 
For spatial binary data (rail joint), 
a rail joint at spatial position $l$ 
is represented as a binary value, where $1$ indicates the rail joint
being positioned there.
Similar to spatiotemporal categorical data,
the one-hot representations $(0,1)$ and $(1,0)$
are used to avoid zero-input.
For each rail joint at spatial position $l$,
FCN of the corresponding data outputs a 4D embedding vector.
The embedding vector is expanded in the temporal and categorical
directions at the expanding layer.
Finally, the set of the expanded embedding vectors is 
concatenated in the spatial direction
to obtain the tensor representation.

For spatiotemporal real number data (ballast age, tonnage, and rainfall),
the embedding layer is not used, and the data are directly fed
into the concatenate layer, because the data are not sparse.

\noindent{\bf ConvLSTM}:
The ConvLSTM computes the spatial/temporal correlation of the track geometry irregularity data and exogenous data.
First, the track geometry irregularity data ${\bm x}_{t-\tau+1},\ldots,{\bm x}_t$ and the exogenous embedding feature ${\bm z}_{t-\tau+1}\ldots {\bm z}_{t}$ are concatenated in the channel dimension by the concatenation layer.
Subsequently, the concatenated data are computed for spatial/temporal correlation using stacked 1D ConvLSTM.
Figure \ref{fig:convlstm_cell} shows the structure of the 1D ConvLSTM cell
with inputs $\mathcal{X}_t$,
hidden states $\mathcal{H}_t$, and cell outputs $\mathcal{C}_t$.
The difference compared with the 2D ConvLSTM cell \cite{shi2015convolutional}
is that the convolution layer is replaced with a 
1D convolution layer. 
The key equations of the 1D ConvLSTM cell are given as follows. 
\begin{eqnarray}
i_t&=&\sigma\left(W_{xi}*\mathcal{X}_t+W_{hi}*\mathcal{H}_{t-1}+W_{ci}\odot \mathcal{C}_{t-1}+b_i\right)\nonumber\\
f_t&=&\sigma\left(W_{xf}*\mathcal{X}_t+W_{hf}*\mathcal{H}_{t-1}+W_{cf}\odot \mathcal{C}_{t-1}+b_f\right)\nonumber\\
\mathcal{C}_t&=&f_t\odot\mathcal{C}_{t-1}+i_t\odot\tanh{\left(W_{xc}*\mathcal{X}_t+W_{hc}*\mathcal{H}_{t-1}+b_c\right)}\label{eq:convlstm_cell}\\
o_t&=&\sigma\left(W_{xo}*\mathcal{X}_t+W_{ho}*\mathcal{H}_{t-1}+W_{co}\odot \mathcal{C}_{t}+b_o\right)\nonumber\\
\mathcal{H}_t&=&o_t\odot\tanh{\left(\mathcal{C}_t\right)}\nonumber,i
\end{eqnarray}
where $*$ denotes the 1D convolution operator, $\odot$ is the Hadamard product,
and $\sigma$ is the sigmoid function. Moreover, $W$ and $b$ are the weight and bias learnable parameters,
respectively.
The output of the ConvLSTM is concatenated in the time dimension by the concatenation layer.
The last convolution layer takes concatenated features as inputs and outputs predictions $\hat{\bm y}_{t+1}$.

\noindent{\bf Loss}:
    The proposed method uses the mean squared error (MSE) loss, which is defined as
    \begin{equation}
        \label{eq:loss}
        \mathcal{L}=\frac{1}{TLC_y}\sum_{t=i}^T\sum_{i=1}^L\sum_{j=1}^{C_y}\left(y_{i,j}^{(t)}-\hat{y}^{(t)}_{i,j}\right)^2
    \end{equation}
    where, 
    $T$ is the number of sequences used for evaluation.

\section*{Experiments}

\subsection*{Experiments setting}
The ConvLSTM is trained for 2,000 epochs, and the parameters with the lowest losses are used on the validation dataset for testing.
The Adam optimizer is also deployed with learning rate $\gamma=0.001$ and betas $(\beta_1,\beta_2)=(0.9,0.999)$.
The ConvLSTM training and all experiments are conducted on a workstation with NVIDIA TITAN RTX (24 GB memory), Intel Core i7-5960X CPU (3.00 GHz), and 64 GB memory.
The ConvLSTM is implemented using Python 3.7.10 and PyTorch 1.8.1.

    \begin{table}[tb]
    \centering
    \begin{tabular}{|l|c|c|c|}
        \hline
        Model & Spatial calculation & Exogenous factor & Nonlinear\\
        \hline
        Linear Regression & &  & \\
        LSTM w/ exogenous & & \checkmark & \checkmark \\
        GRU w/ exogenous & & \checkmark & \checkmark \\
        ConvLSTM w/ exongenous & \checkmark & \checkmark & \checkmark\\
        \hline
    \end{tabular}
    \caption{\label{tab:comparison_methods}Comparison of forecasting methods. Spatial calculation indicates whether the method performs spatial calculations. Exogenous factor indicates whether the method considers exogenous factors. Nonlinear refers to whether the method is nonlinear.}
    \end{table}
    
\subsection*{Evaluation}
The root mean squared error (RMSE),
R-squared ($R^2$), and accuracy are used as evaluation metrics.
RMSE is defined as
\begin{equation}
    \textrm{RMSE}({Y})=\sqrt{\frac{\sum_{(y,\hat{y})\in{Y}}(y-\hat{y})^2}{n({Y})}}
\end{equation}
where $Y\subset \mathbb{R}^2$ is the evaluated dataset, that is, a set of pairs of vertical alignments $y_{i,j}^{(t)}$ and their predictions $\hat{y}_{i,j}^{(t)}$. Moreover, $n(Y)$ is the number of elements in $Y$.
A lower RMSE indicates better forecasting performance
and is helpful for decision making with regards to whether the maintenance operations need to be performed.
$R^2$ is also used to present a clear view of the model performance, which is defined as 
\begin{equation}
    R^2=1-\frac{\sum_{(y,\hat{y})\in{Y}}(y-\hat{y})^2}
    {\sum_{(y,\hat{y})\in{Y}}(y-\bar{y})^2},
\end{equation}
where $\bar{y}=\sum_{(y,\hat{y})\in Y} y/ n(Y)$ is the mean of vertical alignment $y_{i,j}^{(t)}$.
Best possible score of $R^2$ is 1.0.
 In this study, the accuracy shows the percentage of data for which the predictions $\hat{y}$ are within a specific tolerance range $(y-\varepsilon,y+\varepsilon)$ from the observed values $y$. 
 An indicator function $\phi$ is defined that indicates whether the predicted value is within tolerance $\varepsilon$ as follows:
\begin{equation}
    \label{eq:indicator_accuracy}
    \phi(y,\hat{y},\varepsilon)=
    \begin{cases}
        0 & (|y-\hat{y}|\geq\varepsilon)\\
        1 & (|y-\hat{y}|<\varepsilon)
    \end{cases}.
\end{equation}
Using Eq.~(\ref{eq:indicator_accuracy}), the accuracy[\%] is defined as
\begin{equation}
    \textrm{accuracy}(Y,\varepsilon)=\frac{\sum_{(y,\hat{y})\in Y}\phi(y,\hat{y},\varepsilon)}{n(Y)}.
\end{equation}
A higher accuracy indicates a better forecasting performance.

To maintain safety, railroad operators must forecast track geometry.
Therefore, the forecasting model must deliver high performance for data in which vertical alignment is degraded.
The ConvLSTM is evaluated on the entire dataset 
\begin{equation}
    Y=\left\{\left(y_{l,c_y}^{(t)},\hat{y}_{l,c_y}^{(t)}\right)\middle|l=1\ldots L,c_y=1\ldots C_y,t=1\ldots T\right\}
\end{equation}
as well as on the data in which the vertical alignment is less than the threshold level $\alpha$. 
\begin{equation}
    Y=\left\{\left(y_{l,c_y}^{(t)},\hat{y}_{l,c_y}^{(t)}\right)\middle|y_{l,c_y}^{(t)}<\alpha, l=1\ldots L,c_y=1\ldots C_y,t=1\ldots T\right\}.
\end{equation} 
In the experiment, the ConvLSTM is evaluated with $\alpha=-4.0,-6.0$ [mm].

        \begin{table}[t]
        \centering
        \begin{tabular}{|l||r|r|r|r|}
        \hline
        \multirow{2}{*}{Model} & \multicolumn{3}{|c|}{RMSE [mm] ($\downarrow$)} & $R^2$($\uparrow$) \\
        \cline{2-5}
        & Entire & $<-4$ [mm] & $<-6$ [mm] & Entire \\
        \hline \hline
        Linear regression       & 0.502 & 1.196 & 2.477 & 0.798\\
        LSTM\cite{LSTM} w/ exogenous       & 0.302 & 1.091 & 2.411 & 0.927 \\
        GRU\cite{GRU} w/ exogenous        & 0.300 & 1.085 & 2.406 & 0.928 \\
        ConvLSTM w/ exogenous (proposed)   & \textbf{0.293} & \textbf{1.071} & \textbf{2.343} &  \textbf{0.931}\\
        \hline
        \end{tabular}
        \caption{\label{tab:comparison_RMSE}Comparison results of the RMSE and 
        $R^2$. The RMSE is calculated for each method on both the entire data and data with the evaluation threshold levels $\alpha=-4,-6$ [mm].}
        \end{table}
    \begin{table}[t]
    \centering
    \begin{tabular}{|l||r|r|r||r|r|r|}
    \hline
    \multirow{3}{*}{Model} & \multicolumn{6}{|c|}{Accuracy (\%) ($\uparrow$)} \\
    \cline{2-7}
    & \multicolumn{3}{|c||}{$<-4$ [mm]} & \multicolumn{3}{c|}{$<-6$ [mm]} \\
    \cline{2-7}
    & $\pm0.3$ [mm] & $\pm0.5$ [mm] & $\pm1.0$ [mm] & $\pm0.3$ [mm] & $\pm0.5$ [mm] & $\pm1.0$ [mm] \\
    \hline \hline
    Linear regression       & 63.96 & 75.78 & 85.06 & \textbf{33.59} & \textbf{44.66} & \textbf{58.84} \\
    LSTM w/ exogenous       & 56.55 & 72.51 & 85.13 & 05.44 & 17.67 & 46.02 \\
    GRU w/ exogenous        & 61.21 & 74.96 & 86.01 & 16.12 & 30.87 & 49.71 \\
    ConvLSTM w/ exogenous (proposed)  & \textbf{66.48} & \textbf{77.82} & \textbf{87.35} & 26.02 & 37.28 & 54.76 \\
    \hline
    \end{tabular}
    \caption{\label{tab:comparison_accuracy}Comparison results of the accuracy. The accuracy is calculated for each method with tolerance $\varepsilon=0.3,0.5,1.0$ [mm] on the data with the evaluation threshold levels $\alpha=-4,-6$ [mm].}
    \end{table}

\subsection*{Comparison experiment}
    In the comparison experiments, the proposed method is compared with other forecasting methods.
    The curves of the losses for training and validation data
    are detailed in Supplementary Figure S1 online.
    After learning and computing each method, the evaluation metrics and output series are evaluated.
    Table \ref{tab:comparison_methods} provides a preliminary comparison of multiple forecasting methods.
    For LSTM and GRU, we determine each architecture by tuning
    the number of layers (see Supplementary Section S.4).
    The details of the compared methods are discussed next.

    \noindent{\bf Linear regression}:
        Linear regression is the simplest baseline method.
        Because vertical alignments vary nonlinearly in a time series, linear regression is performed with a sliding window.
        The linear model is defined as $\hat{y}_{t+1}=b_t x_{t+1}+a_t$, 
        where $y_{t+1}$ and $x_{t+1}$ are the vertical alignment and inspection dates, respectively, of the $(t+1)$-th inspection.
        Moreover, $a_t$ and $b_t$ are the parameters of this method, which are optimized using the least-squares method for the last three vertical alignments $\{y_{t-2},y_{t-1},y_t\}$ and inspection dates $\{x_{t-2},x_{t-1},x_t\}$.
        In contrast to the proposed method, this method neither considers exogenous factors nor performs spatial calculations.
        
    \noindent{\bf LSTM with exogenous data}:
        LSTM~\cite{LSTM} is a variant of recurrent neural networks (RNNs). 
        RNNs, including the LSTM and GRU, are used for natural language processing and time-series forecasting.
        This method replaces the ConvLSTM with the LSTM in the proposed model architecture.
        In contrast to the proposed method, this method does not involve spatial computations, but does consider the exogenous factors.
        The hyperparameters of the LSTM follow Pytorch examples
        (For tuning the number of layers,
        see Supplementary Section S.4.).
        The learning setting of this method is similar to that of the proposed method.
        
    \noindent{\bf GRU with exogenous data}:
        Similar to LSTM, GRU~\cite{GRU} is a variant of RNNs.
        This method replaces the ConvLSTM with a GRU in the proposed model architecture.
        In contrast to the proposed method, this method does not involve spatial computations, but does consider the exogenous factors.
        The hyperparameters of the GRU also follow Pytorch examples.
        (For tuning the number of layers,
        see Supplementary Section S.4.).
        The learning setting of this method is also similar to that of the proposed method.

\begin{table}[t]
\centering
\begin{tabular}{|l|r|}
\hline
Model & Training time (s) \\
\hline \hline
Linear regression       & - \\
LSTM w/ exogenous       & 9293 \\ 
GRU w/ exogenous        & 6875 \\ 
ConvLSTM w/ exogenous (proposed)  & 26353\\ 
\hline
\end{tabular}
\caption{\label{tab:training_time} Training time for each model.}
\end{table}

\subsection*{Ablation study on exogenous factors}
    In an ablation study on exogenous factors, 
     the significance of these factors is verified with regards to forecasting.
    Specifically, the evaluation metrics with and without the input of specific exogenous data are compared. 
    If the difference in the evaluation metrics is significant, the exogenous data are critical for forecasting using the proposed method. 
    The proposed method is trained and evaluated under the following conditions:
    \begin{itemize}
        \item A case that inputs all exogenous data.
        \item A case that inputs all exogenous data except a specific one.
        \item A case that does not input all exogenous data. (This case is vanilla ConvLSTM.)
    \end{itemize}

\begin{figure}
    \centering
    \subfloat[]{\includegraphics[width=\linewidth]{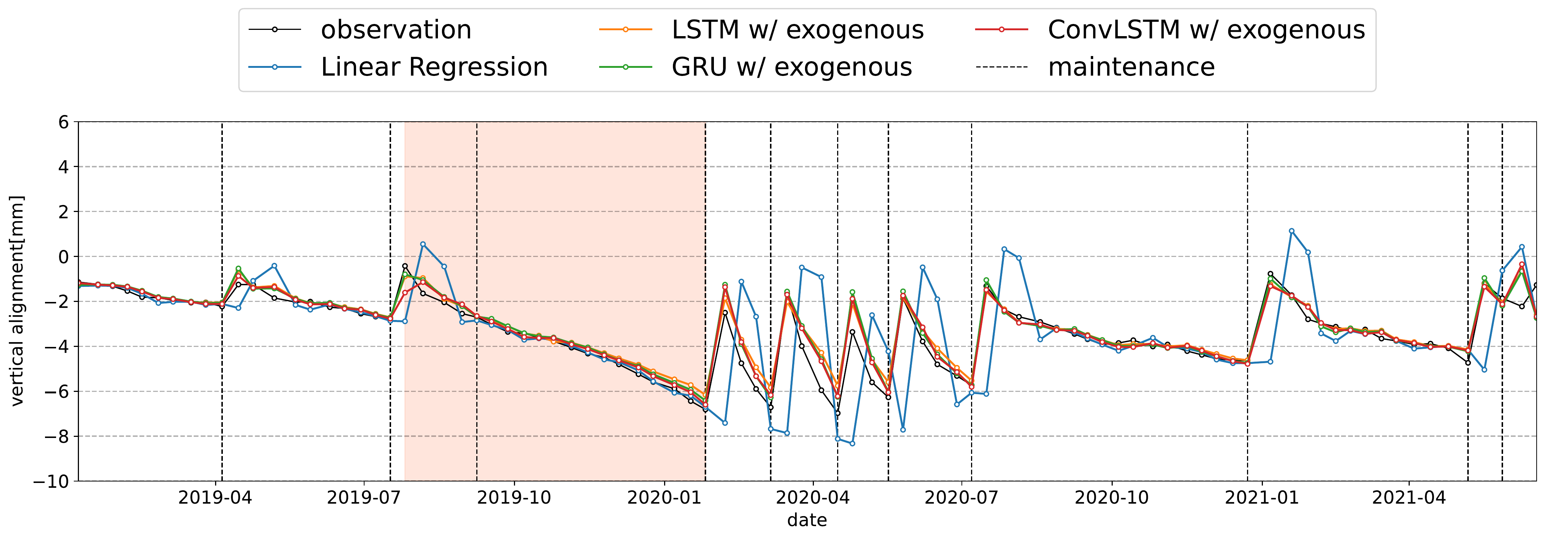}\label{fig:compare_wave1}}
    
    \subfloat[]{\includegraphics[width=\linewidth]{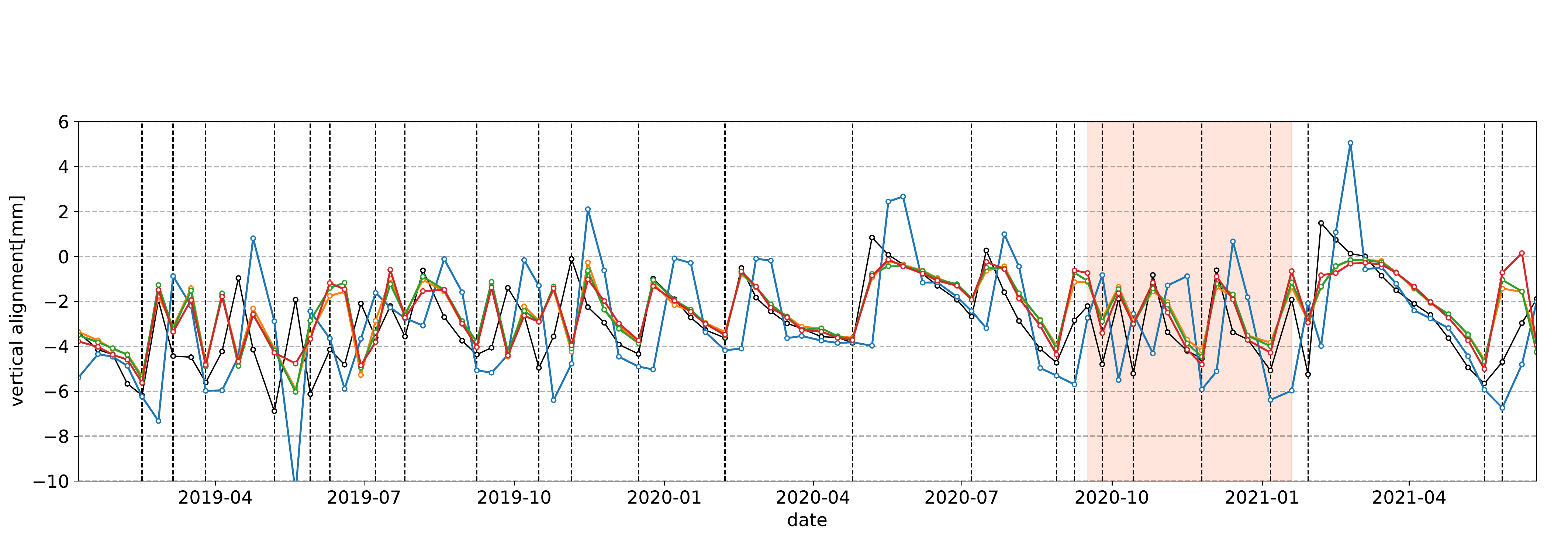}\label{fig:compare_wave3}}
    
    \centering
    \subfloat[]{\includegraphics[width=0.45\linewidth]{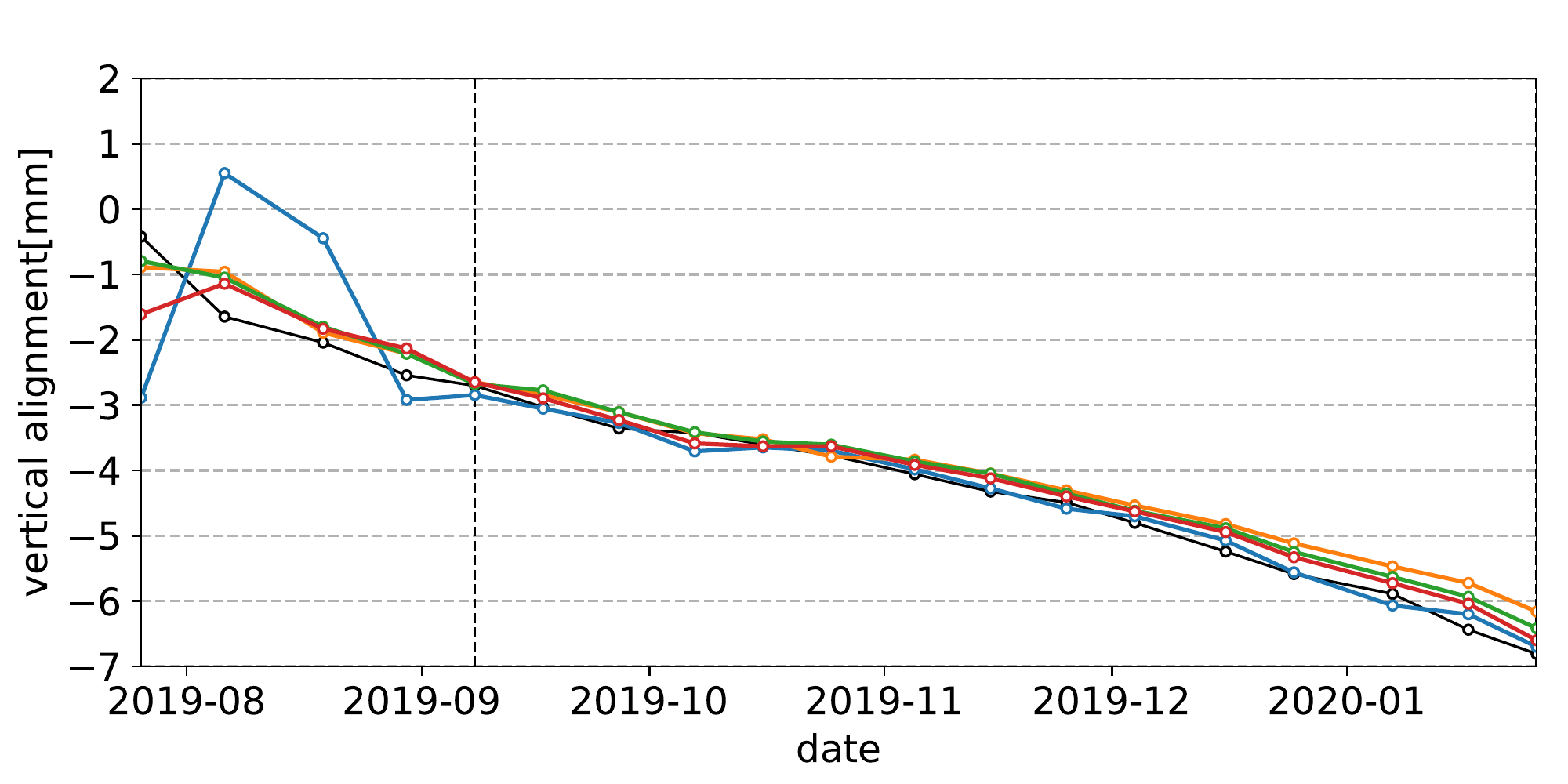}\label{fig:compare_wave2}} \quad
    \subfloat[]{\includegraphics[width=0.45\linewidth]{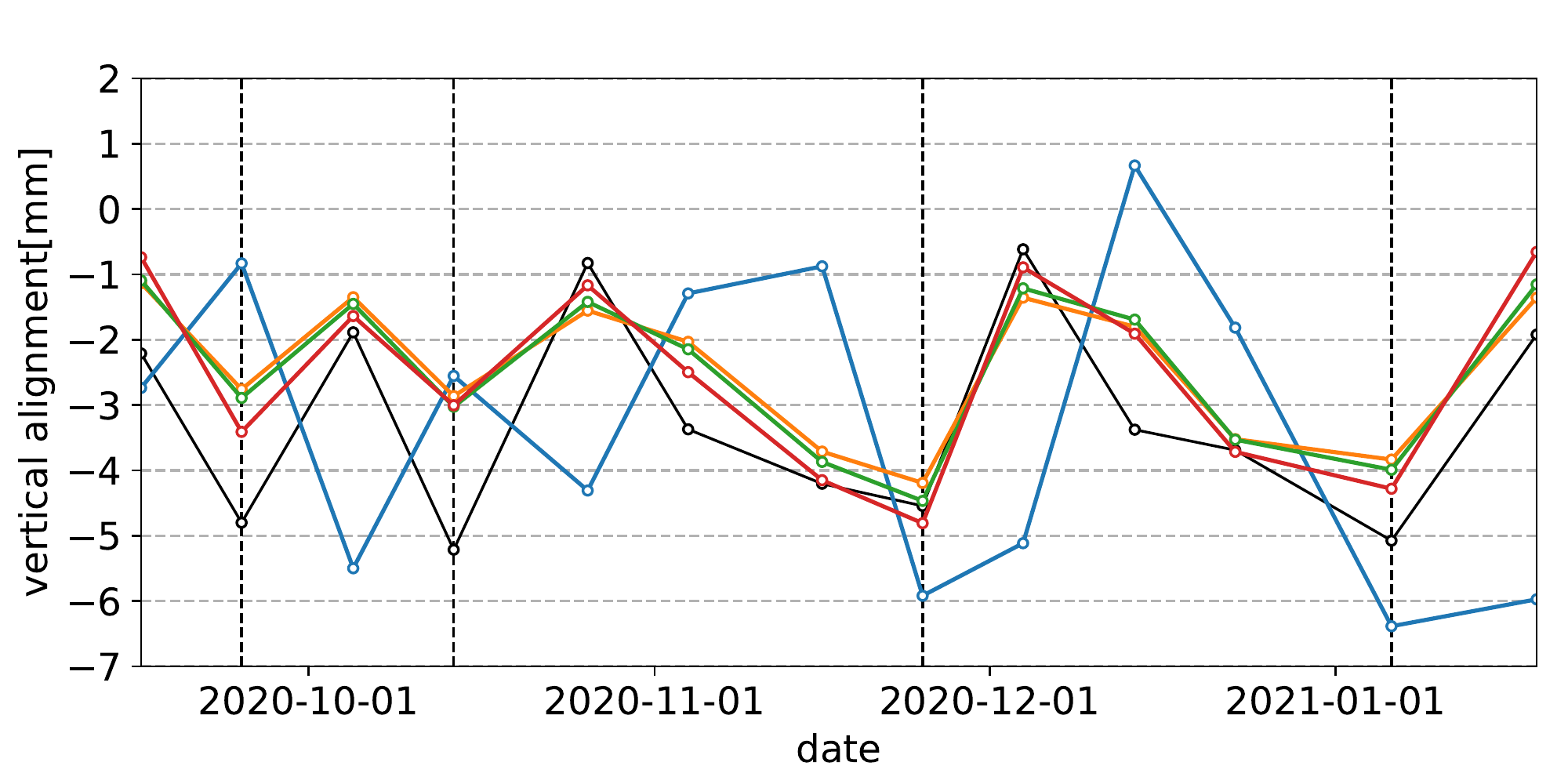}\label{fig:compare_wave4}}
    \caption[]{Forecasted time series of specific spatial points for each method. 
    The vertical dotted lines indicate the dates when maintenance operations were performed.
    \subref{fig:compare_wave1} is the time series of a specific spatial point with a normal maintenance frequency. \subref{fig:compare_wave2} is the orange part of \subref{fig:compare_wave1}. \subref{fig:compare_wave3} is the time series of a specific spatial point with a high maintenance frequency. \subref{fig:compare_wave4} is the orange part of \subref{fig:compare_wave3}.
    }
\label{fig:compare_wave}
\end{figure}

\section*{Results and discussion}
\subsection*{Comparison results}
Tables \ref{tab:comparison_RMSE} and \ref{tab:comparison_accuracy} summarize the comparison results of the RMSE, $R^2$, and accuracy, respectively, between the proposed method and other forecasting methods.
The proposed method achieves the lowest RMSE for both the entire dataset and the dataset with thresholds $\alpha=-4,-6$ [mm] 
and the highest $R^2$ for the entire dataset.
    This result indicates that spatial calculations
    improve the forecasting performance in terms of RMSE.
Moreover, the proposed method has the highest accuracy for the data threshold $\alpha=-4$ [mm],
whereas the simplest baseline, linear regression, has the highest accuracy for the data threshold $\alpha=-6$ [mm].
This result indicates that the vertical alignment varies linearly when the vertical alignment is less than $\alpha=-6$ [mm].
 However, linear regression demonstrates a poorer performance at points with a high maintenance frequency (See Discussion).

Figure \ref{fig:compare_wave} shows the forecasted time series of the spatial points on the track for each method. 
Note that the vertical dotted lines indicate the dates
when maintenance operations were performed.
Figure \ref{fig:compare_wave1} and~\ref{fig:compare_wave2} show the forecasted time series at the spatial point where maintenance is performed once every eight inspections on average, that is, approximately once every three months.
The spatial exogenous data at the spatial point
are as follows: 
the under-structure is excavation and no rail joint exists.
Each method showcases a high forecasting performance at the spatial point with a normal maintenance frequency.
When the vertical alignment varies linearly, as shown in Figure \ref{fig:compare_wave2}, linear regression performs better than the proposed method.
In contrast, for a relatively high maintenance frequency, such as once a month, e.g., from January to July 2020, in Figure \ref{fig:compare_wave1}, linear regression forecasts indicate a poorer performance compared with other methods.

Figure \ref{fig:compare_wave3} and~\ref{fig:compare_wave4} show the forecasted time series at the spatial point where maintenance is performed once every four inspections on average, that is, approximately once every month. 
The spatial exogenous data at the spatial point
are as follows: 
the under-structure is embankment and no rail joint exists.
Linear regression performs worse than the other methods shown in Figure \ref{fig:compare_wave3} and~\ref{fig:compare_wave4},
and provides the predictions with a delayed interval of one inspection.
This delay is due to the fact that linear regression does not use the maintenance records, but only the past data.
In Figure \ref{fig:compare_wave4}, the proposed method exhibits a higher performance than the LSTM and GRU. This is because the ConvLSTM in the proposed method captures the spatial correlation by convolution, whereas the LSTM and GRU do not.

Table \ref{tab:training_time} lists the training time for each model.
For linear regression, the training time 
is not provided because it is too short.
For LSTM and GRU, the training time is 
the cumulative training time at all spatial points.
As listed in Table \ref{tab:training_time},
ConvLSTM takes a longer training time than the other methods.
However, the training time for ConvLSTM is not significantly long
because the task involves 10-day ahead forecasting.

\begin{figure}
    \centering
    \subfloat[]{\includegraphics[width=0.5 \linewidth]{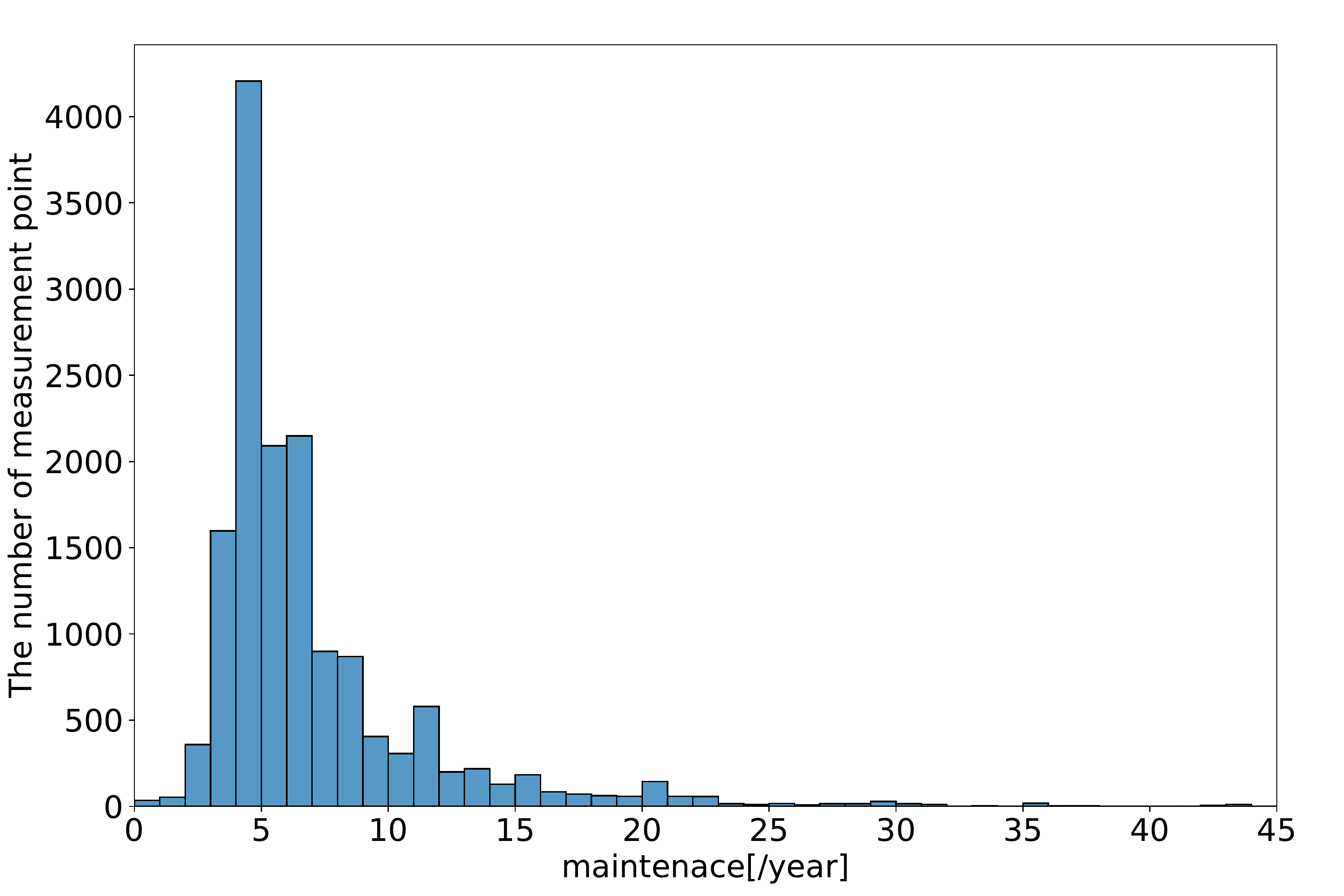}\label{fig:hist}}
    \subfloat[]{\includegraphics[width=0.5 \linewidth]{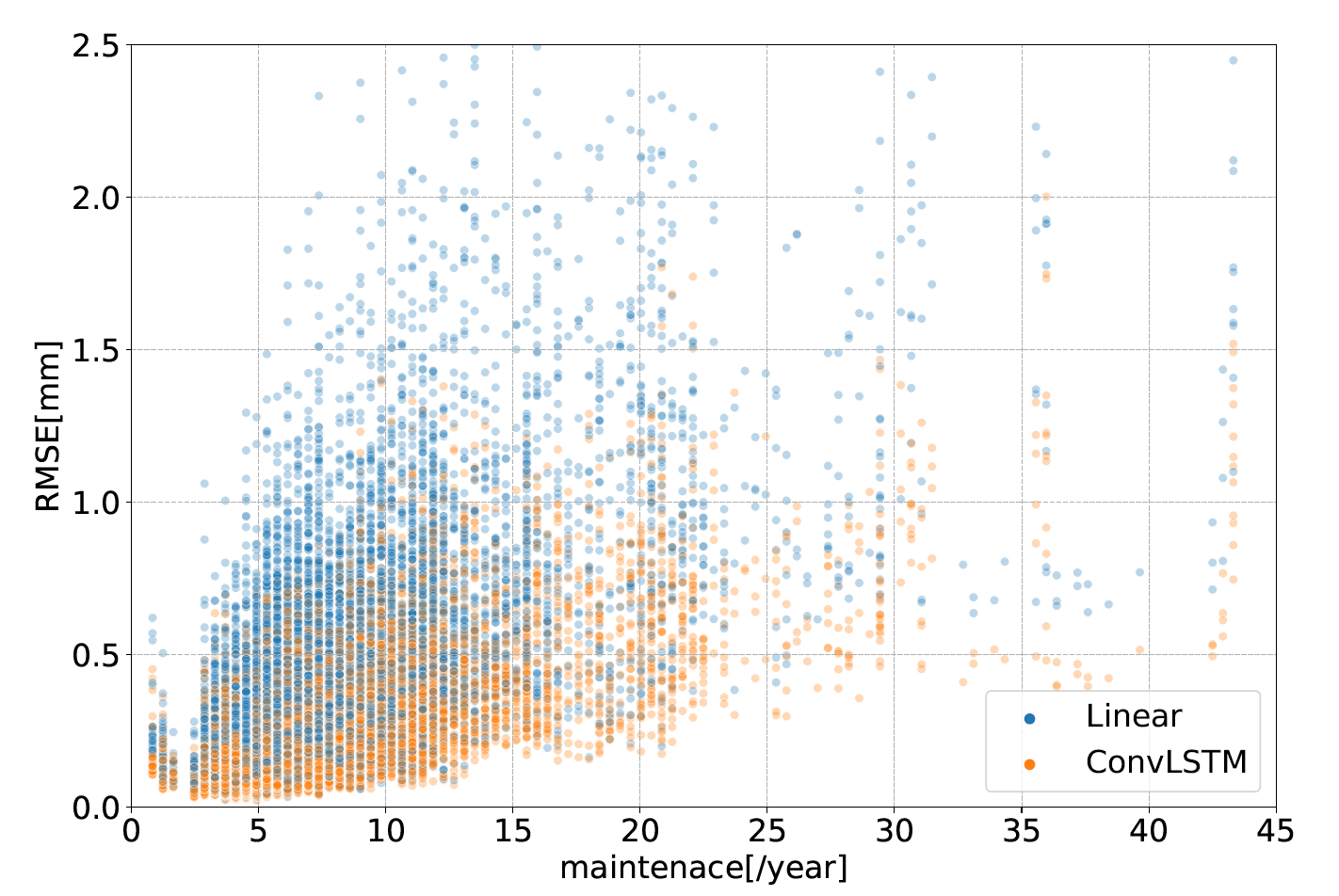}\label{fig:scatter2}}
\caption[]{\subref{fig:hist} Histogram of the number of maintenances at each spatial point on the rail. \subref{fig:scatter2} Scatter plot of the number of track maintenances and RMSEs at each measurement point.}
\label{fig:hist and scatter}
\end{figure}

\subsection*{Discussion}
Herein, the limitation of the linear regression model and the effectiveness of the proposed ConvLSTM are discussed.
For many spatiotemporal points,
the linear regression forecast is evidently sufficient because the vertical alignment varies linearly, as shown in Figure \ref{fig:compare_wave1} and \ref{fig:compare_wave2}.
However, linear regression performs poorly at points with a high maintenance frequency, as shown in Figure \ref{fig:compare_wave3}.
Next,
the proposed method is compared with the linear regression model at points with a high maintenance frequency.
Figure \ref{fig:hist} shows a histogram of the average maintenance frequency per year for each measurement point in the test data.
At most points, the maintenance frequency per year is approximately five.
Figure \ref{fig:scatter2} shows a scatter plot  of the track maintenance frequency and RMSEs at each measurement point.
With a higher maintenance frequency, the ConvLSTM tends to have a lower RMSE than the linear regression.
The measurement interval of the track geometry irregularity data is approximately 10 days.
Therefore, if maintenance is performed 12 times annually, the track geometry irregularities are measured an average of three times during the two maintenances.
A prior research has fitted a multi-stage linear model during two maintenances ~\cite{chang2010multi}.
However, vertical alignments are difficult to forecast with such a linear model because of the limited data available when track geometry irregularities are only inspected three or four times during two maintenances, that is, when the maintenance frequency is high.
In contrast, the ConvLSTM can achieve a high forecasting performance even at spatial points with a high maintenance frequency.

\begin{table}[t]
{\footnotesize
\centering
\begin{tabular}{|l|c|c|c|c|c|c||r|r|r|}
\hline
\multirow{2}{*}{Case name} &\multicolumn{6}{|c||}{Exogenous data} & \multicolumn{3}{|c|}{RMSE [mm] ($\downarrow$)} \\
\cline{2-10}
 & Maintenance & Structure & Rail joint & Ballast age & Tonnage & Rainfall & Entire & $<-4$ [mm] & $<-6$ [mm] \\
\hline \hline
w/ all&\checkmark & \checkmark & \checkmark & \checkmark & \checkmark & \checkmark & \textbf{0.293} & \textbf{1.071} & \textbf{2.343} \\ 
\hdashline
w/o maintenance & & \checkmark & \checkmark & \checkmark & \checkmark & \checkmark & 0.361 & 1.405 & 3.056 \\ 
w/o structure & \checkmark &  & \checkmark & \checkmark & \checkmark & \checkmark & 0.294 & 1.075 & 2.361 \\
w/o rail joint & \checkmark & \checkmark &  & \checkmark & \checkmark & \checkmark & 0.294 & 1.078 & 2.369 \\ 
w/o ballast age & \checkmark & \checkmark & \checkmark &  & \checkmark & \checkmark & 0.294 & 1.083 & 2.370 \\ 
w/o tonnage & \checkmark & \checkmark & \checkmark & \checkmark &  & \checkmark & \textbf{0.293} & 1.076 & 2.360 \\ 
w/o rainfall & \checkmark & \checkmark & \checkmark & \checkmark & \checkmark &  & 0.294 & 1.072 & 2.354 \\ 
\hdashline
w/o all&  &  &  &  &  &  & 0.360 & 1.303 & 2.883 \\ 
\hline
\end{tabular}
\caption{\label{tab:ablation_RMSE}Results of the ablation study (RMSE). The RMSE is calculated for each method using both the entire data and the data with the threshold levels $\alpha=-4,-6$ [mm]. }
}
\end{table}

\begin{table}[t]
{\footnotesize
\centering
\begin{tabular}{|l|c|c|c|c|c|c||r|r|r|}
\hline
\multirow{3}{*}{Case name} & \multicolumn{6}{|c||}{Exogenous data} & \multicolumn{3}{|c|}{Accuracy (\%)  ($\uparrow$)} \\
\cline{2-10}
& \multirow{2}{*}{Maintenance} & \multirow{2}{*}{Structure} & \multirow{2}{*}{Rail joint} & \multirow{2}{*}{Ballast age} & \multirow{2}{*}{Tonnage} & \multirow{2}{*}{Rainfall} &  \multicolumn{3}{|c|}{$<-4$ [mm]} \\
\cline{8-10}
& & & & & & & $\pm 0.3$ [mm] & $\pm 0.5$ [mm] & $\pm 1.0$ [mm] \\
\hline \hline
w/ all &\checkmark & \checkmark & \checkmark & \checkmark & \checkmark & \checkmark & 66.48 & 77.82 & \textbf{87.35} \\
\hdashline
 w/o maintenance & & \checkmark & \checkmark & \checkmark & \checkmark & \checkmark & 15.04 & 29.17 & 58.32 \\
w/o structure & \checkmark &  & \checkmark & \checkmark & \checkmark & \checkmark & 65.54 & 77.53 & 86.83 \\
w/o rail joint & \checkmark & \checkmark &  & \checkmark & \checkmark & \checkmark & 66.29 & 77.79 & 87.06 \\
w/o ballast age & \checkmark & \checkmark & \checkmark &  & \checkmark & \checkmark & 65.61 & 77.56 & 87.03 \\
w/o tonnage & \checkmark & \checkmark & \checkmark & \checkmark &  & \checkmark & 66.40 & 77.75 & 86.95 \\
w/o rainfall & \checkmark & \checkmark & \checkmark & \checkmark & \checkmark &  & \textbf{66.83} & \textbf{78.16} & 87.17 \\
\hdashline
 w/o all & &  &  &  &  &  & 17.96 & 31.65 & 62.78 \\
\hline
\end{tabular}
\caption{\label{tab:ablation_accuracy_4mm}Results of the ablation study (accuracy). The accuracy is calculated for each method with tolerance $\varepsilon=0.3,0.5,1.0$ [mm] using data with the threshold level $\alpha=-4$ [mm].}
}
\end{table}
\begin{table}[t]
{\footnotesize
\centering
\begin{tabular}{|l|c|c|c|c|c|c||r|r|r|}
\hline
\multirow{3}{*}{Case name} & \multicolumn{6}{|c||}{Exogenous data} & \multicolumn{3}{|c|}{Accuracy (\%)  ($\uparrow$)} \\
\cline{2-10}
& \multirow{2}{*}{Maintenance} & \multirow{2}{*}{Structure} & \multirow{2}{*}{Rail joint} & \multirow{2}{*}{Ballast age} & \multirow{2}{*}{Tonnage} & \multirow{2}{*}{Rainfall} &  \multicolumn{3}{|c|}{$<-6$mm} \\
\cline{8-10}
& & & & & & & $\pm 0.3$ [mm] & $\pm 0.5$ [mm] & $\pm 1.0$ [mm] \\
\hline \hline
w/ all & \checkmark & \checkmark & \checkmark & \checkmark & \checkmark & \checkmark & 26.02 & 37.28 & \textbf{54.76}\\
\hdashline
w/o maintenance & & \checkmark & \checkmark & \checkmark & \checkmark & \checkmark & 00.00 & 00.00 & 00.39 \\
w/o structure & \checkmark &  & \checkmark & \checkmark & \checkmark & \checkmark & 25.83 & 37.09 & 52.43 \\
w/o rail joint & \checkmark & \checkmark &  & \checkmark & \checkmark & \checkmark & 22.14 & 35.92 & 52.82 \\
w/o ballast age & \checkmark & \checkmark & \checkmark &  & \checkmark & \checkmark & 23.88 & 37.86 & 54.17 \\
w/o tonnage & \checkmark & \checkmark & \checkmark & \checkmark &  & \checkmark & 24.66 & \textbf{38.83} & 53.98 \\
w/o rainfall & \checkmark & \checkmark & \checkmark & \checkmark & \checkmark &  & \textbf{27.57} & 38.45 & 54.17 \\
\hdashline
w/o all& &  &  &  &  &  & 00.00 & 00.00 & 00.58 \\
\hline
\end{tabular}
\caption{\label{tab:ablation_accuracy_6mm}Results of the ablation study (accuracy). The accuracy is calculated for each method with tolerance $\varepsilon=0.3,0.5,1.0$ [mm] using the data with threshold level $\alpha=-6$ [mm].}
}
\end{table}

\subsection*{Ablation study on exogenous factors}
Tables \ref{tab:ablation_RMSE},~\ref{tab:ablation_accuracy_4mm}, and~\ref{tab:ablation_accuracy_6mm} summarize the results of the ablation study based on the exogenous factors for the ConvLSTM.
For the ablation studies of the LSTM and GRU,
please see Supplementary Tables S2, S3, and S4 (online).
In the tables,
“With-all” refers to a case in which all exogenous data are used. 
"Without specific data" refers to the case in which all exogenous data except specific data are used. 
For example, “without-maintenance” refers to the case that uses all exogenous data except for track maintenance data. 
“Without-all” refers to the case in which no exogenous factor is used.
Comparing “with-all” and “without-maintenance”, “without-maintenance” shows a higher RMSE for both the entire dataset and the dataset with thresholds $\alpha=-4,-6$ [mm].
Moreover, “without-maintenance” shows a lower accuracy than “with-all” for the dataset with thresholds $\alpha=-4,-6$ [mm].
The difference between the “with-all” and “without-maintenance” evaluation metrics is significant, as maintenance data are useful for forecasting in the proposed method.
However, for the other exogenous data, the evaluation metrics do not worsen when the exogenous data are removed. 
Furthermore, the RMSE and accuracy are also not sensitive 
even in the vicinity of
the rail joints and the boundary between
two under-structures, such as embankment and excavation.
Therefore, whether other exogenous data improve the forecasting performance of the proposed method cannot be confirmed.
Note that this result does not imply that other exogenous factors
are not causally related to vertical track alignment.
For example, tonnage is related to track geometry irregularities \cite{guler2014prediction},
the effect may be hidden.
The time index implicitly contains the information
about the tonnage, because the tonnage is the same along the 15 km track portion.

\begin{figure}
    \centering
    \subfloat[]{\includegraphics[width=\linewidth]{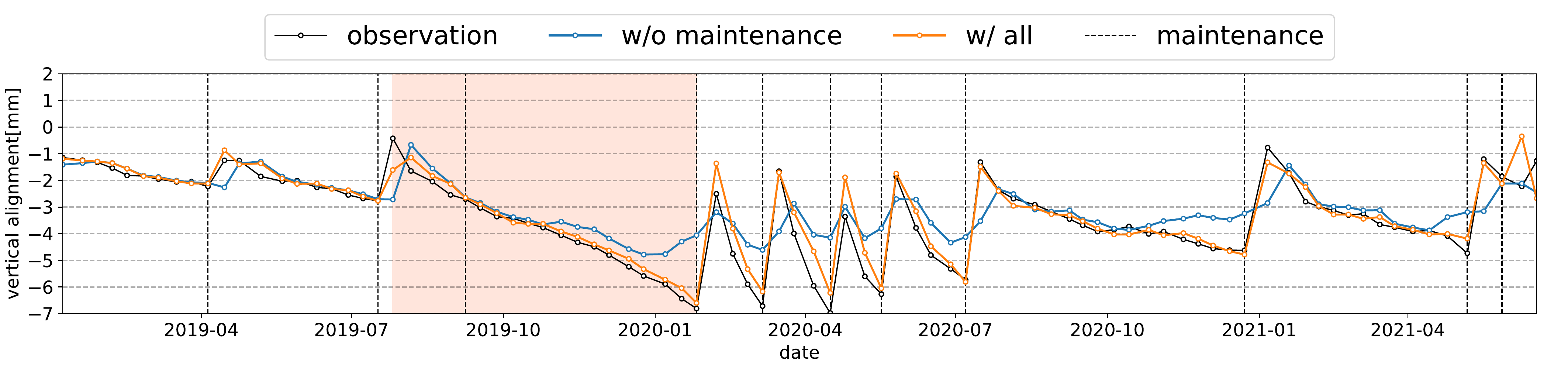}\label{fig:ablation_wave1}}
    
    \subfloat[]{\includegraphics[width=\linewidth]{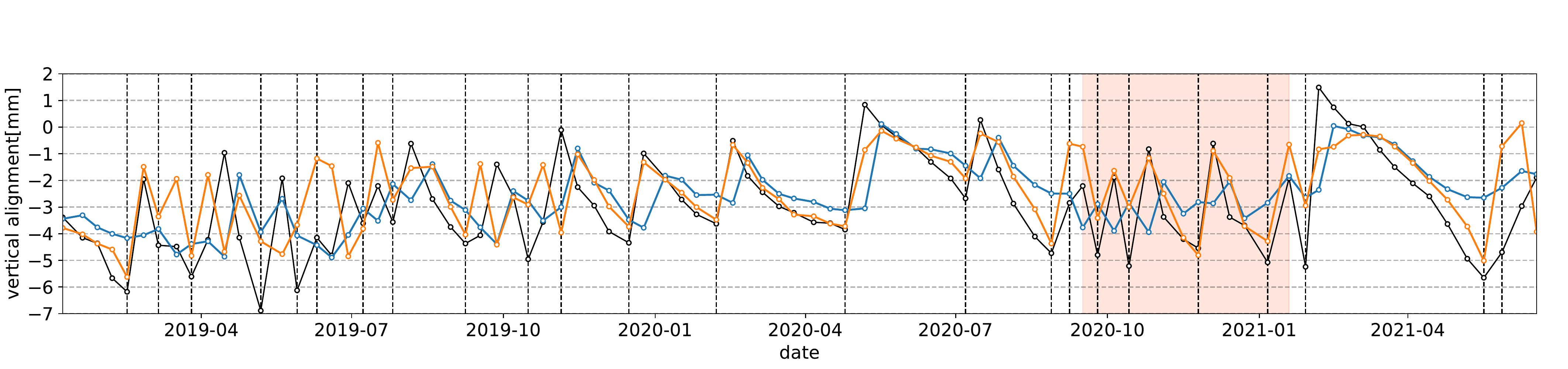}\label{fig:ablation_wave3}}
    
    \centering
    \subfloat[]{\includegraphics[width=0.45\linewidth]{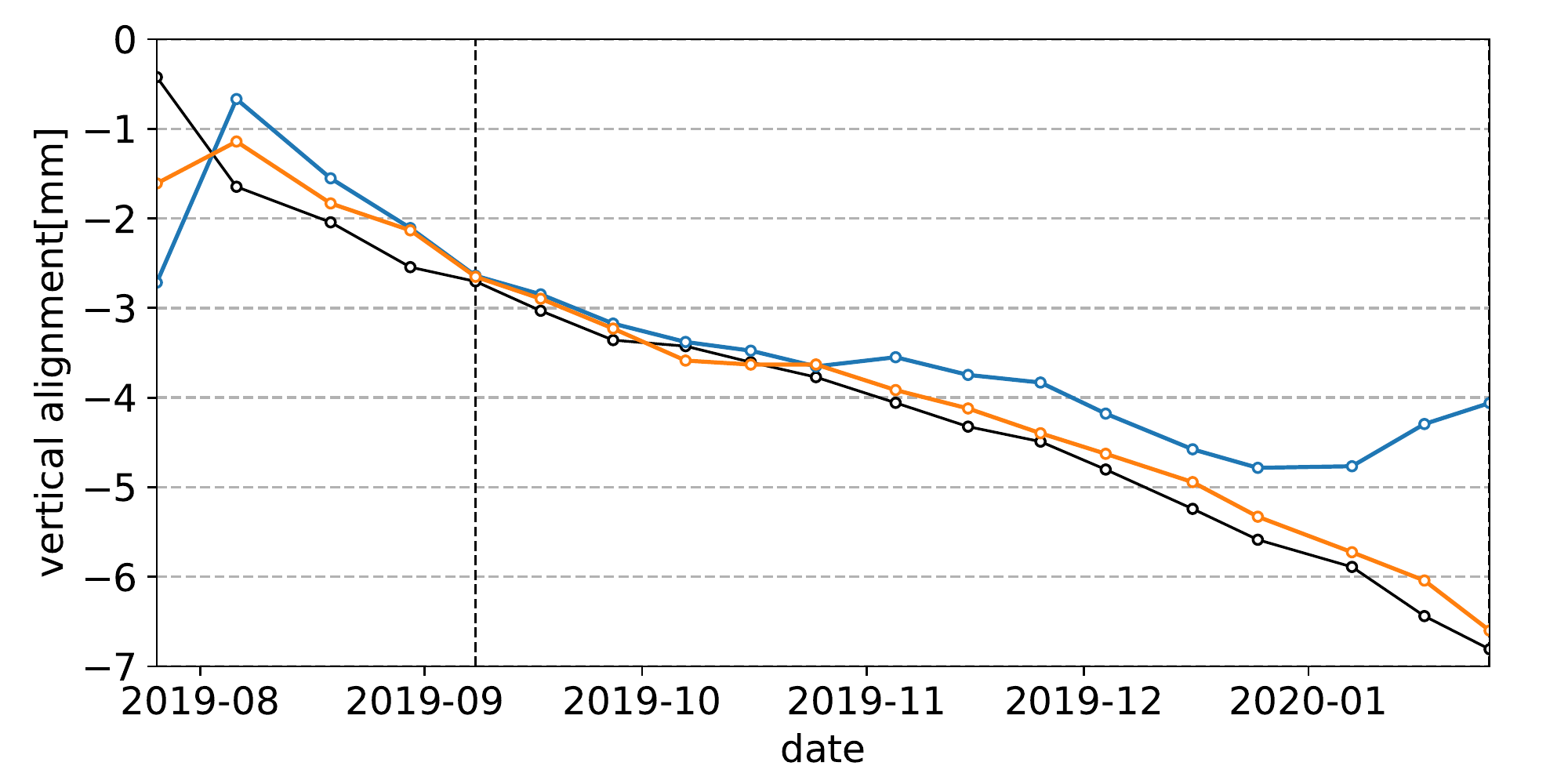}\label{fig:ablation_wave2}} \quad
    \subfloat[]{\includegraphics[width=0.45\linewidth]{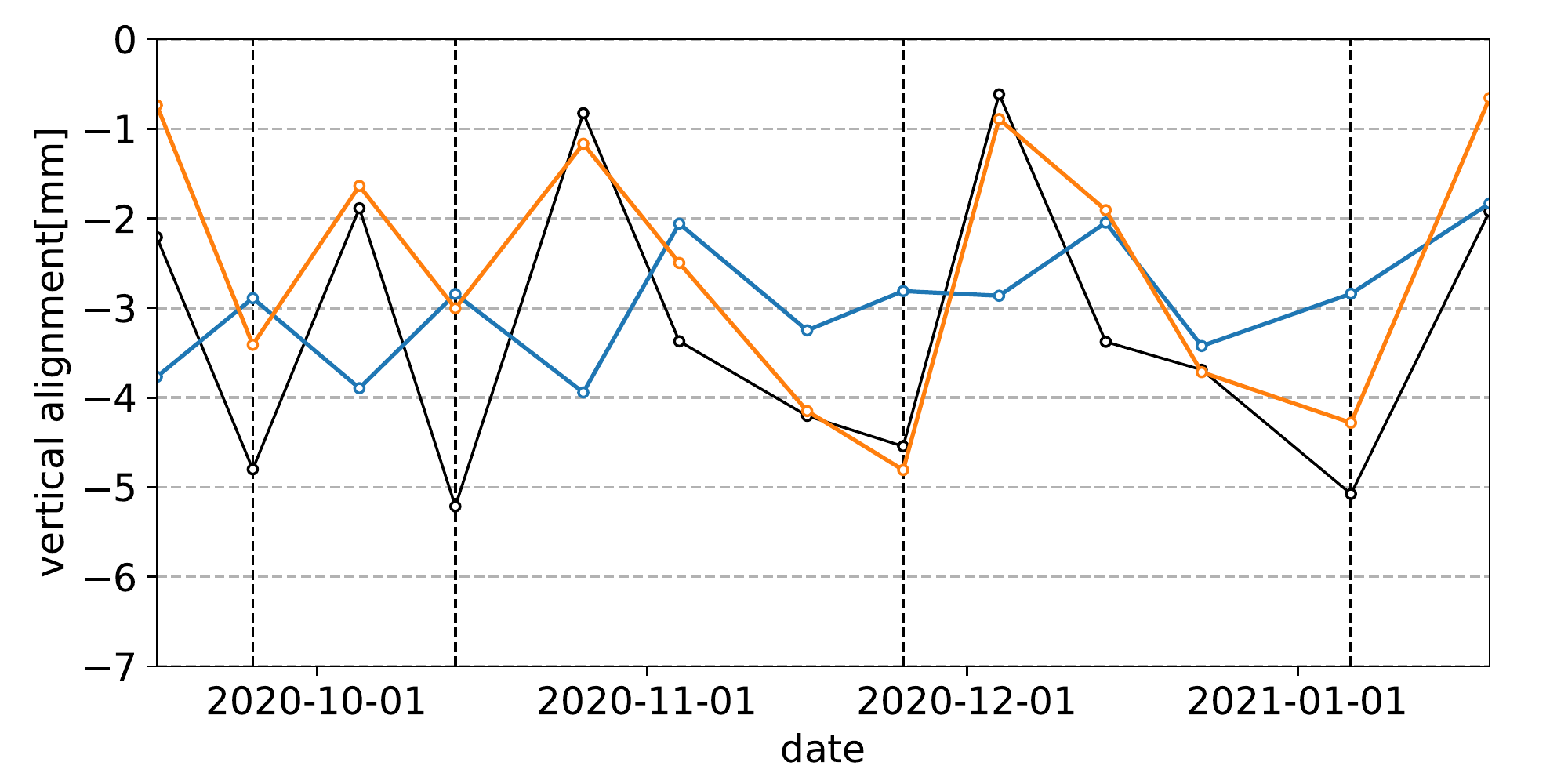}\label{fig:ablation_wave4}}
    \caption[]{Forecasted time series of specific spatial points under “with-all” and “without-maintenance” conditions. The vertical dotted lines indicate the dates when maintenance operations were performed.
    \subref{fig:ablation_wave1} is the time series of a specific spatial point with a normal maintenance frequency. \subref{fig:ablation_wave2} is the orange part of \subref{fig:ablation_wave1}. \subref{fig:ablation_wave3} is the time series of a specific spatial point with a high maintenance frequency. \subref{fig:ablation_wave4} is the orange part of \subref{fig:ablation_wave3}.
    }
\label{fig:ablation_wave}
\end{figure}

Figure \ref{fig:ablation_wave} shows the forecasted time series 
with and without maintenance
at the same spatial points as in Figure \ref{fig:compare_wave}.
Figure \ref{fig:ablation_wave1} and~\ref{fig:ablation_wave2} show the forecasted time series of a specific spatial point where maintenance is performed once every eight inspections on average, that is, approximately once every three months.
The “with-all” condition showcases a higher performance than the “without-maintenance” condition.
This is because in the “without-maintenance” condition it is not known when an increase in vertical alignment occurs due to maintenance.
The lower the vertical alignment, the more intermediate is the forecasting between the increase owing to maintenance and the decrease owing to degradation.

Figure \ref{fig:ablation_wave3} and~\ref{fig:ablation_wave4} show the forecasted time series of a specific spatial point where maintenance is performed once every four inspections on average, that is, approximately once every month.
At a spatial point with a high maintenance frequency, the forecasted time series of the “with-all” condition is typically closer to the observed time series than that of the “without-maintenance” condition.

\section*{Conclusion}
This study proposes a ConvLSTM to forecast vertical alignments of railroad at high spatial and temporal frequencies. 
The ConvLSTM is designed to consider spatial correlations and the effect of exogenous factors on vertical alignment.
The proposed method is experimentally compared with other methods in terms of the forecasting performance. 
Additionally, an ablation study on exogenous factors is conducted to examine their contribution to the forecasting performance. 
The results reveal that spatial calculations and maintenance record data improve the forecasting of the vertical alignment.
Additionally, the proposed method showcases a higher forecasting performance at points with a high maintenance frequency.

The ConvLSTM has the potential to be applied to other tracks, although its applicability should be carefully considered in the case that the exogenous data for the tracks are very different from our experimental data. For example, linear regression may be sufficient if maintenance frequency is low. However, as the experiment showed, spatial calculation and maintenance records improve forecast performance when maintenance operations are frequently required significantly.
Furthermore, the explainability of the ConvLSTM is of practical significance and should be pursued in future work, because the explainable ConvLSTM will provide insights into which exogenous factors are useful for forecasting track geometry irregularities.

\section*{Acknowledgments}
This study was funded by the Central Japan Railway Company.

%Bibliography
\bibliographystyle{unsrt}  
\bibliography{sample}  

\appendix
\def\thesection{Appendix \Alph{section}}

\newpage
\section{Maintenance operations for vertical alignment correction}
    We list the nine maintenance operations used for
    forecasting vertical track alignment in Table \ref{tab:maintenance_list}
    of the main manuscript.
    These nine operations are categorized by merging more detailed operations.
    Here we show the full list of the maintenance operations including
    the detailed operations in Table \ref{tab:maintenace_list}.
    For example, 
    the category of sleeper maintenance includes sleeper replacement, loose sleeper repair, and so on. 

    \begin{table}[ht]
    \centering
    \begin{tabular}{|l|l|}
        \hline
        Detailed maintenance & Merged category \\
        \hline
        Uneven fixing  & Uneven fixing \\
        \hdashline
        Tamping by multiple tie tamper & Tamping by multiple tie tamper \\
        \hdashline
        Manual tamping & Manual tamping\\
        \hdashline
        Ballast replacement  & Ballast replacement \\
        \hdashline
        Right rail replacement  & Right rail replacement \\
        \hdashline
        Left rail replacement  & Left rail replacement \\
        \hdashline
        Sleeper replacement & \multirow{6}{*}{Sleeper maintenance } \\
        Sleeper installation  & \\
        Loose sleeper repair  & \\
        Sleeper alignment  & \\
        Sleeper removal  & \\
        Sleeper relocation  & \\
        \hdashline
        Remediation of mud-pumping & Remediation of mud-pumping \\
        \hdashline
        Expansion joint replacement  & \multirow{4}{*}{Others} \\
        Turnout replacement  & \\
        Glued insulation joint replacement  & \\
        Rail grinding  & \\
        \hline
    \end{tabular}
    \caption{\label{tab:maintenace_list} Full list of maintenance operations for vertical alignment correction}
    \end{table}

\section{Loss curves for training and validation data}
\begin{figure}[ht]
    \centering
    \subfloat[ConvLSTM]{\includegraphics[width=0.31\linewidth]{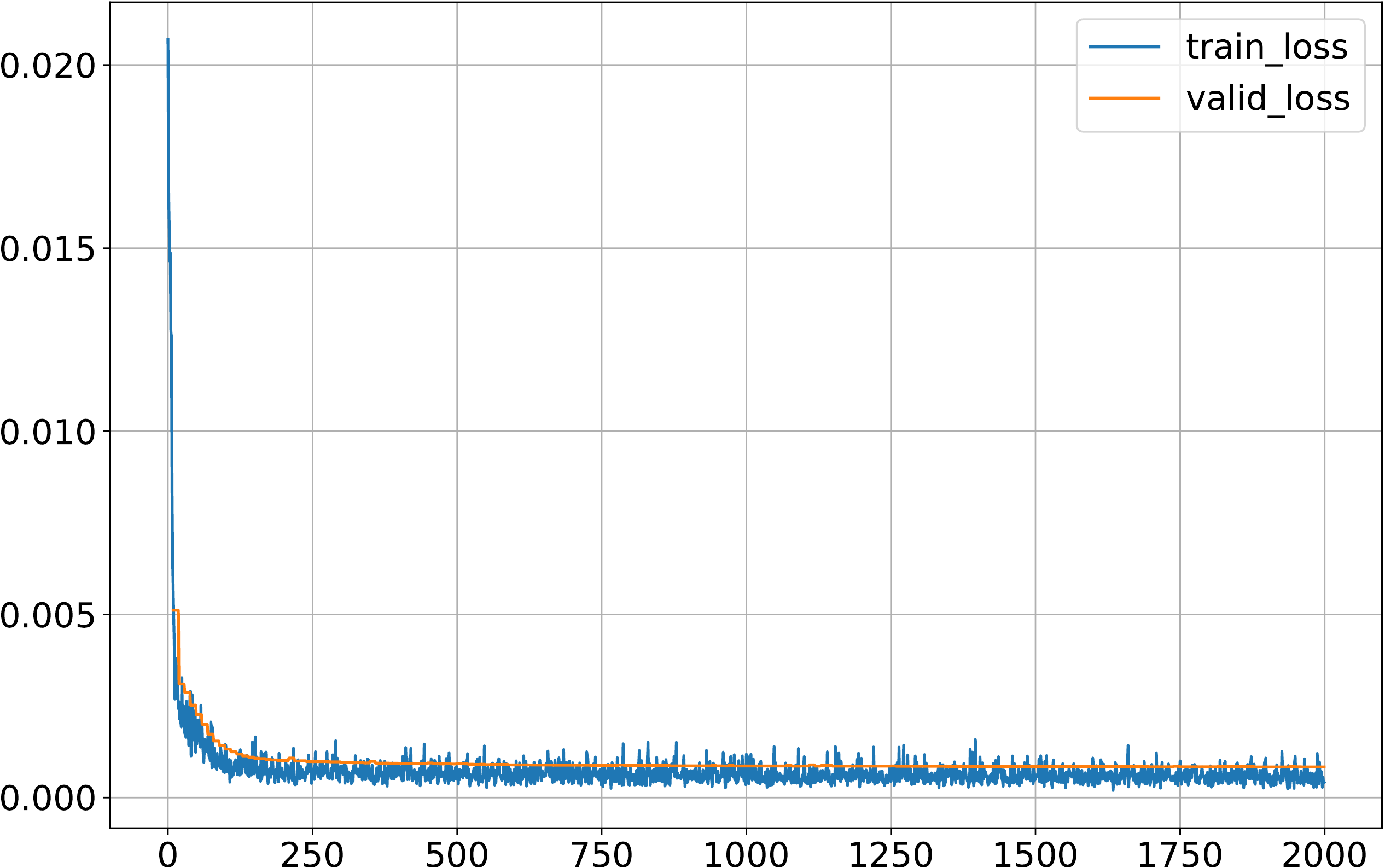}\label{fig:convlstm_loss}}\quad
    \subfloat[GRU]{\includegraphics[width=0.31\linewidth]{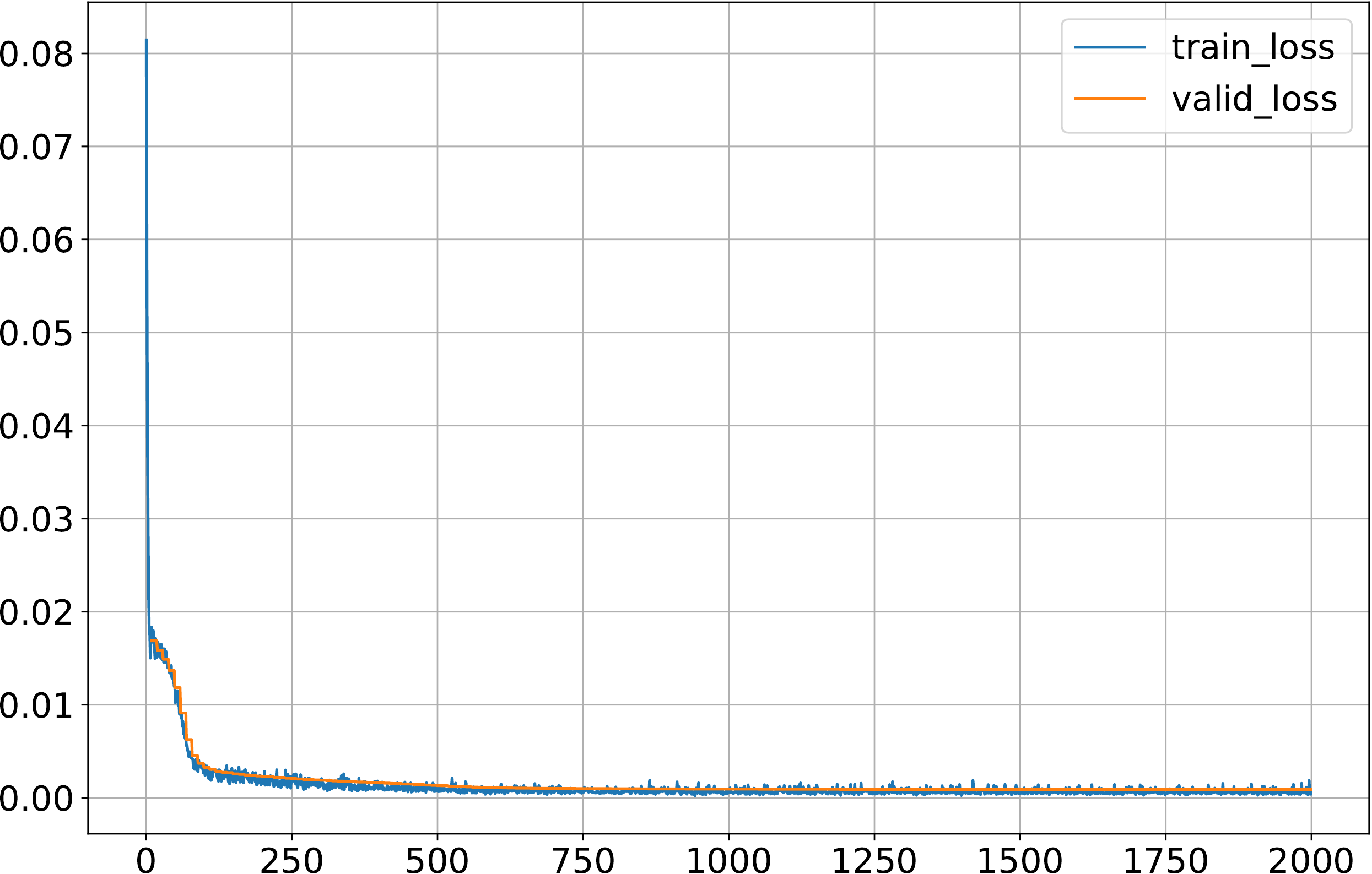}\label{fig:gru_loss}}\quad
    \subfloat[LSTM]{\includegraphics[width=0.31\linewidth]{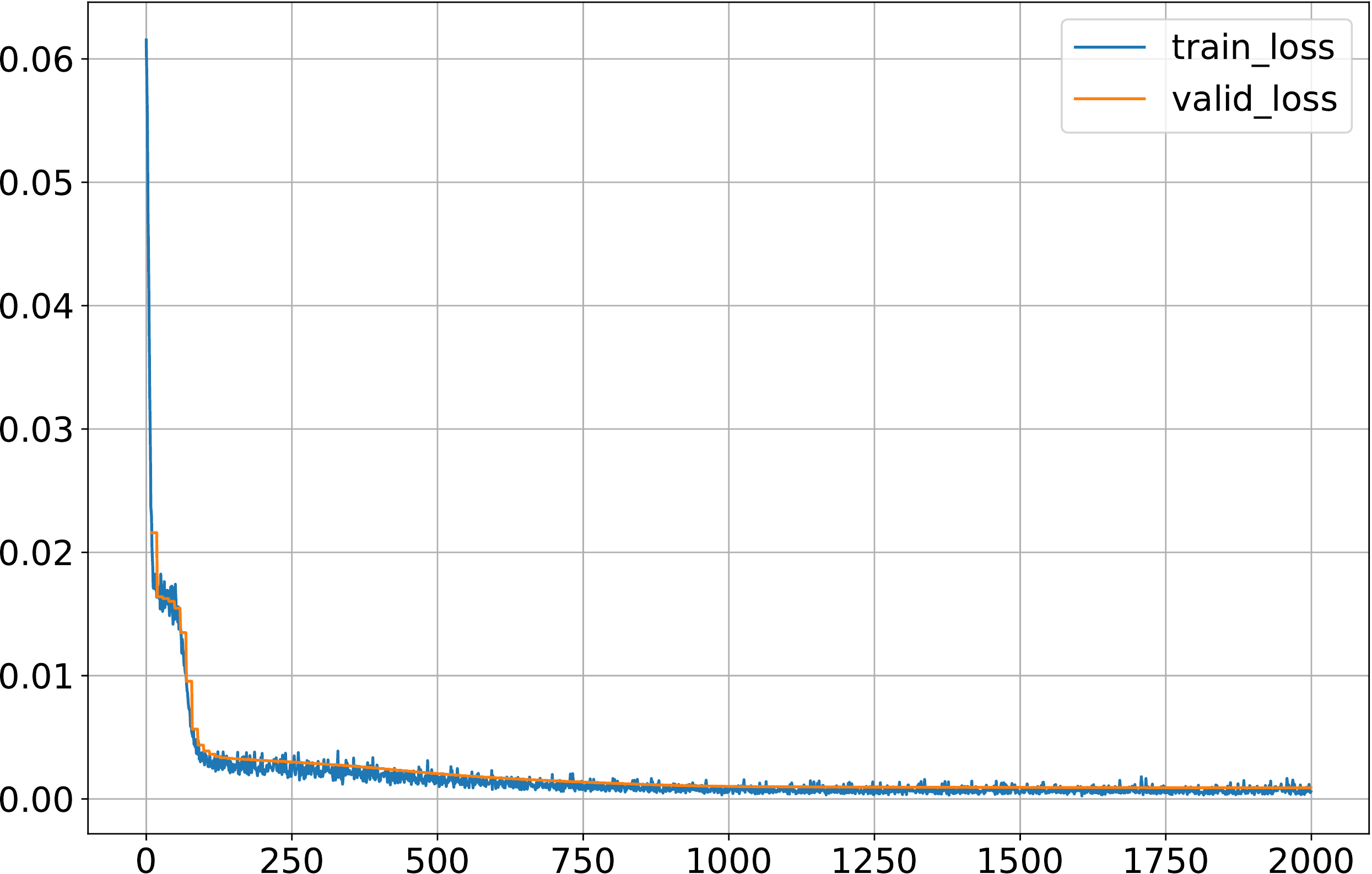}\label{fig:lstm_loss}}
    \caption{The loss curves of (a) ConvLSTM, (b) GRU, and (c) LSTM.}
    \label{fig:loss}
    \end{figure}

Figure \ref{fig:loss} show the loss curves of 
ConvLSTM, GRU, and LSTM in the comparison experiment.
The blue and orange lines show the losses for training and validation data, respectively.
As shown in Figure \ref{fig:loss},
the loss curves for both the training and validation data decrease as the epoch progresses.
These results indicate overfitting does not occur for all
ConvLSTM, GRU, and LSTM.

%\newpage
\section{Ablation study on exogenous factors for LSTM and GRU}
    Tables \ref{tab:ablation_rmse}, \ref{tab:ablation_accuracy_4mm_2} and \ref{tab:ablation_accuracy_6mm_2} 
    show the results of the ablation study on the exogenous factor
    for LSTM and GRU.
    The results are similar to those of ConvLSTM
    (see Tables 9, 10, and 11 in the main manuscript).
    For example, comparing “with-all” and “without-maintenance”, 
    “without-maintenance” shows a higher RMSE and a lower accuracy
    for both the entire dataset and the dataset with thresholds
    $\alpha = -4, -6$ [mm].
    Therefore, the maintenance records are also significant for forecasting in LSTM and GRU. 
    Additionally, the results prove that ConvLSTM outperforms
    LSTM and GRU  (see Tables \ref{tab:ablation_RMSE}, \ref{tab:ablation_accuracy_4mm}, and \ref{tab:ablation_accuracy_6mm} in the main manuscript).

    \begin{table}[htb]
    {\footnotesize
        \centering
        \subfloat[LSTM]{\label{tab:ablation_LSTM_RMSE}
            \begin{tabular}{|l|c|c|c|c|c|c||r|r|r|}
\hline
\multirow{2}{*}{Case name} &\multicolumn{6}{|c||}{Exogenous data} & \multicolumn{3}{|c|}{RMSE(mm) ($\downarrow$)} \\
\cline{2-10}
 & Maintenance & Structure & Rail joint & Ballast age & Tonnage & Rainfall & Entire & $<-4$mm & $<-6$mm \\
\hline \hline
w/ all&\checkmark & \checkmark & \checkmark & \checkmark & \checkmark & \checkmark & 0.302 & 1.091 & 2.477 \\ 
\hdashline
w/o maintenance & & \checkmark & \checkmark & \checkmark & \checkmark & \checkmark & 0.369 & 1.226 & 2.651 \\ 
w/o structure & \checkmark &  & \checkmark & \checkmark & \checkmark & \checkmark & 0.305 & 1.114 & 2.526 \\
w/o rail joint & \checkmark & \checkmark &  & \checkmark & \checkmark & \checkmark & 0.302 & 1.104 & 2.469 \\ 
w/o ballast age & \checkmark & \checkmark & \checkmark &  & \checkmark & \checkmark & 0.305 & 1.091 & 0.247 \\ 
w/o tonnage & \checkmark & \checkmark & \checkmark & \checkmark &  & \checkmark & 0.305 & 1.090 & 2.417 \\ 
w/o rainfall & \checkmark & \checkmark & \checkmark & \checkmark & \checkmark &  & 0.300 & 1.079 & 2.402 \\ 
\hdashline
w/o all&  &  &  &  &  &  & 0.369 & 1.194 & 2.582 \\ 
\hline
\end{tabular}
    }\\
     \subfloat[GRU]{
\begin{tabular}{|l|c|c|c|c|c|c||r|r|r|}
\hline
\multirow{2}{*}{Case name} &\multicolumn{6}{|c||}{Exogenous data} & \multicolumn{3}{|c|}{RMSE(mm) ($\downarrow$)} \\
\cline{2-10}
 & Maintenance & Structure & Rail joint & Ballast age & Tonnage & Rainfall & Entire & $<-4$mm & $<-6$mm \\
\hline \hline
w/ all&\checkmark & \checkmark & \checkmark & \checkmark & \checkmark & \checkmark & 0.300 & 1.085 & 2.406 \\ 
\hdashline
w/o maintenance & & \checkmark & \checkmark & \checkmark & \checkmark & \checkmark & 0.366  & 1.270 & 2.762 \\ 
w/o structure & \checkmark &  & \checkmark & \checkmark & \checkmark & \checkmark & 0.299 & 1.091 & 2.416 \\
w/o rail joint & \checkmark & \checkmark &  & \checkmark & \checkmark & \checkmark & 0.299 & 1.083 & 2.398 \\ 
w/o ballast age & \checkmark & \checkmark & \checkmark &  & \checkmark & \checkmark & 0.299 & 1.093 & 2.414 \\ 
w/o tonnage & \checkmark & \checkmark & \checkmark & \checkmark &  & \checkmark & 0.299 & 1,089 & 2.402 \\ 
w/o rainfall & \checkmark & \checkmark & \checkmark & \checkmark & \checkmark &  & 0.299 & 1.093 & 2.437 \\ 
\hdashline
w/o all&  &  &  &  &  &  & 0.365 & 1.285 & 2.828 \\ 
\hline
\end{tabular}
    }
    \caption{\label{tab:ablation_rmse}Results of the ablation study (RMSE) for (a) LSTM and (b) GRU. The RMSE is calculated with both the entire data and data with the threshold levels $\alpha=-4,-6$[mm].}
  }
    \end{table}
    %\end{landscape}

%    \begin{landscape}
    \begin{table}[htb]
    {\footnotesize
        \centering
        \subfloat[LSTM]{\label{tab:ablation_LSTM_4mm}
            \begin{tabular}{|l|c|c|c|c|c|c||r|r|r|}
        \hline
        \multirow{3}{*}{Case name} & \multicolumn{6}{|c||}{Exogenous data} & \multicolumn{3}{|c|}{Accuracy(\%)  ($\uparrow$)} \\
        \cline{2-10}
        & \multirow{2}{*}{Maintenance} & \multirow{2}{*}{Structure} & \multirow{2}{*}{Rail joint} & \multirow{2}{*}{Ballast age} & \multirow{2}{*}{Tonnage} & \multirow{2}{*}{Rainfall} &  \multicolumn{3}{|c|}{$<-4$mm} \\
        \cline{8-10}
        & & & & & & & $\pm 0.3$mm & $\pm 0.5$mm & $\pm 1.0$mm \\
        \hline \hline
        w/ all & \checkmark & \checkmark & \checkmark & \checkmark & \checkmark & \checkmark & 56.55 & 72.51 & 85.13 \\
        \hdashline
        w/o maintenance & & \checkmark & \checkmark & \checkmark & \checkmark & \checkmark & 05.14 & 22.74 & 70.60 \\
        w/o structure & \checkmark &  & \checkmark & \checkmark & \checkmark & \checkmark & 57.26 & 71.65 & 84.07\\
        w/o rail joint & \checkmark & \checkmark &  & \checkmark & \checkmark & \checkmark & 57.05 & 72.38 & 84.70\\
        w/o ballast age & \checkmark & \checkmark & \checkmark &  & \checkmark & \checkmark
        & 59.51 & 73.16 & 85.03 \\
        w/o tonnage & \checkmark & \checkmark & \checkmark & \checkmark &  & \checkmark & 59.98 & 73.48 & 85.12 \\
        w/o rainfall & \checkmark & \checkmark & \checkmark & \checkmark & \checkmark &  & 63.66 & 75.83 & 85.83 \\
        \hdashline
        w/o all& &  &  &  &  &  & 05.52 & 25.08 & 73.19\\
        \hline
        \end{tabular}
    }\\
     \subfloat[GRU]{
         \begin{tabular}{|l|c|c|c|c|c|c||r|r|r|}
        \hline
        \multirow{3}{*}{Case name} & \multicolumn{6}{|c||}{Exogenous data} & \multicolumn{3}{|c|}{Accuracy(\%)  ($\uparrow$)} \\
        \cline{2-10}
        & \multirow{2}{*}{Maintenance} & \multirow{2}{*}{Structure} & \multirow{2}{*}{Rail joint} & \multirow{2}{*}{Ballast age} & \multirow{2}{*}{Tonnage} & \multirow{2}{*}{Rainfall} &  \multicolumn{3}{|c|}{$<-4$mm} \\
        \cline{8-10}
        & & & & & & & $\pm 0.3$mm & $\pm 0.5$mm & $\pm 1.0$mm \\
        \hline \hline
        w/ all & \checkmark & \checkmark & \checkmark & \checkmark & \checkmark & \checkmark & 61.21 & 74.96 & 86.01\\
        \hdashline
        w/o maintenance & & \checkmark & \checkmark & \checkmark & \checkmark & \checkmark & 07.56 & 23.32 & 64.36 \\
        w/o structure & \checkmark &  & \checkmark & \checkmark & \checkmark & \checkmark & 63.55 & 76.22 & 86.01 \\
        w/o rail joint & \checkmark & \checkmark &  & \checkmark & \checkmark & \checkmark & 63.40 & 76.11 & 85.97 \\
        w/o ballast age & \checkmark & \checkmark & \checkmark &  & \checkmark & \checkmark & 63.66 & 75.83 & 85.71 \\
        w/o tonnage & \checkmark & \checkmark & \checkmark & \checkmark &  & \checkmark & 63.83 & 76.08 & 85.76 \\
        w/o rainfall & \checkmark & \checkmark & \checkmark & \checkmark & \checkmark &  & 64.48 & 76.59 & 86.26 \\
        \hdashline
        w/o all& &  &  &  &  &  & 09.24 & 24.70 & 64.02 \\
        \hline
        
        \end{tabular}
    }
    \caption{\label{tab:ablation_accuracy_4mm_2}Results of the ablation study (accuracy) for (a) LSTM and (b) GRU. The accuracy is calculated with tolerance $\varepsilon=0.3,0.5,1.0$[mm] on the data with the evaluation threshold levels $\alpha=-4$[mm].}
    }
    \end{table}
%    \end{landscape}

%\newpage
%    \begin{landscape}
    \begin{table}[htb]
    {\footnotesize
        \centering
        \subfloat[LSTM]{\label{tab:ablation_LSTM_accuracy_6mm}
            \begin{tabular}{|l|c|c|c|c|c|c||r|r|r|}
        \hline
        \multirow{3}{*}{Case name} & \multicolumn{6}{|c||}{Exogenous data} & \multicolumn{3}{|c|}{Accuracy(\%)  ($\uparrow$)} \\
        \cline{2-10}
        & \multirow{2}{*}{Maintenance} & \multirow{2}{*}{Structure} & \multirow{2}{*}{Rail joint} & \multirow{2}{*}{Ballast age} & \multirow{2}{*}{Tonnage} & \multirow{2}{*}{Rainfall} &  \multicolumn{3}{|c|}{$<-6$mm} \\
        \cline{8-10}
        & & & & & & & $\pm 0.3$mm & $\pm 0.5$mm & $\pm 1.0$mm \\
        \hline \hline
        w/ all & \checkmark & \checkmark & \checkmark & \checkmark & \checkmark & \checkmark & 05.44 & 17.67 & 46.02\\
        \hdashline
        w/o maintenance & & \checkmark & \checkmark & \checkmark & \checkmark & \checkmark & 00.00 & 00.00 & 01.36 \\
        w/o structure & \checkmark &  & \checkmark & \checkmark & \checkmark & \checkmark & 06.21 & 21.75 & 42.52 \\
        w/o rail joint & \checkmark & \checkmark &  & \checkmark & \checkmark & \checkmark & 22.14 & 35.92 & 52.82 \\
        w/o ballast age & \checkmark & \checkmark & \checkmark &  & \checkmark & \checkmark & 04.66 & 20.58 & 45.05 \\
        w/o tonnage & \checkmark & \checkmark & \checkmark & \checkmark &  & \checkmark & 09.71 & 24.66 & 46.02 \\
        w/o rainfall & \checkmark & \checkmark & \checkmark & \checkmark & \checkmark &  & 15.15 & 30.10 & 48.74 \\
        \hdashline
        w/o all& &  &  &  &  &  & 00.00 & 00.00 & 03.30 \\
        \hline
        \end{tabular}
    }\\
     \subfloat[GRU]{
         \begin{tabular}{|l|c|c|c|c|c|c||r|r|r|}
        \hline
        \multirow{3}{*}{Case name} & \multicolumn{6}{|c||}{Exogenous data} & \multicolumn{3}{|c|}{Accuracy(\%)  ($\uparrow$)} \\
        \cline{2-10}
        & \multirow{2}{*}{Maintenance} & \multirow{2}{*}{Structure} & \multirow{2}{*}{Rail joint} & \multirow{2}{*}{Ballast age} & \multirow{2}{*}{Tonnage} & \multirow{2}{*}{Rainfall} &  \multicolumn{3}{|c|}{$<-6$mm} \\
        \cline{8-10}
        & & & & & & & $\pm 0.3$mm & $\pm 0.5$mm & $\pm 1.0$mm \\
        \hline \hline
        w/ all & \checkmark & \checkmark & \checkmark & \checkmark & \checkmark & \checkmark & 16.12 & 30.87 & 49.71\\
        \hdashline
        w/o maintenance & & \checkmark & \checkmark & \checkmark & \checkmark & \checkmark & 00.00 & 00.00 & 01.55 \\
        w/o structure & \checkmark &  & \checkmark & \checkmark & \checkmark & \checkmark & 24.08 & 36.50 & 51.84 \\
        w/o rail joint & \checkmark & \checkmark &  & \checkmark & \checkmark & \checkmark & 23.88 & 35.34 & 51.65 \\
        w/o ballast age & \checkmark & \checkmark & \checkmark &  & \checkmark & \checkmark & 21.36 & 33.20 & 50.29 \\
        w/o tonnage & \checkmark & \checkmark & \checkmark & \checkmark &  & \checkmark & 22.33 & 34.17 & 51.07 \\
        w/o rainfall & \checkmark & \checkmark & \checkmark & \checkmark & \checkmark &  & 20.97 & 34.17 & 51.46 \\
        \hdashline
        w/o all& &  &  &  &  &  & 00.00 & 00.00 & 01.55 \\
        \hline
        \end{tabular}
    }
    \caption{\label{tab:ablation_accuracy_6mm_2}Results of the ablation study (accuracy) for (a) LSTM and (b) GRU. The accuracy is calculated with tolerance $\varepsilon=0.3,0.5,1.0$[mm] on the data with the evaluation threshold levels $\alpha=-6$[mm].}
    }
    \end{table}
%    \end{landscape}

%    \begin{landscape}

\newpage
\section{Layer tuning of LSTM and GRU}
    To determine the best architectures for LSTM and GRU, we examine
    the forecasting performance by changing
    the number of layers for LSTM and GRU.
    Tables \ref{tab:hypara_RMSE} and \ref{tab:hypara_accuracy}
    show the RMSE and accuracy of the tuning results, respectively.
    In the tables, the number of layers means that the architecture
    consists of that number of layers.
    We also show the results by ConvLSTM for comparison.
    In each case, ConvLSTM provides better RMSE and accuracy 
    than those obtained by tuning LSTM and GRU.
    
        \begin{table}[htb]
        \centering
    \subfloat[LSTM]{\begin{tabular}{|l||r|r|r|}
       \hline
    \multirow{2}{*}{Num. layers} & \multicolumn{3}{|c|}{RMSE(mm) ($\downarrow$)} \\
    \cline{2-4}
    & Entire & $<-4$mm & $<-6$mm \\
    \hline \hline
1          & 0.300 & 1.096 & 2.451  \\
2          & 0.302 & 1.091 & 2.411  \\
3          & 0.304 & 1.107 & 2.460 \\
4          & 0.315 & 1.137 & 2.538 \\\hline
ConvLSTM   & 0.293 & 1.071 & 2.343 \\
    \hline
    \end{tabular}\label{tab:hypara_LSTM_RMSE}
    }
     %   \caption{Tuning number of layers for LSTM}    &
    \subfloat[GRU]{
    \begin{tabular}{|l||r|r|r|}
       \hline
    \multirow{2}{*}{Num. layers} & \multicolumn{3}{|c|}{RMSE(mm) ($\downarrow$)} \\
    \cline{2-4}
    & entire & $<-4$mm & $<-6$mm \\
    \hline \hline
    1          & 0.300& 1.097 & 2.428  \\
    2          & 0.300& 1.085 & 2.406  \\
    3          & {0.299}& 1.090 & 2.412  \\
    4          & {0.299}& {1.083} & {2.399}  \\\hline
    ConvLSTM   & 0.293& 1.071 & 2.343  \\
    \hline
    \end{tabular}
    }
        \caption{\label{tab:hypara_RMSE} RMSE results by tuning the number of layers for (a) LSTM and (b) GRU.}
    \end{table}
    
    \begin{table}[htb]
        \centering
    \subfloat[LSTM]{\begin{tabular}{|l||r|r|r|r|r|r|}
     \hline
     \multirow{3}{*}{Num. layers} & \multicolumn{6}{|c|}{Accuracy(\%) ($\uparrow$)} \\
    \cline{2-7}
    & \multicolumn{3}{|c|}{$<-4$mm} & \multicolumn{3}{c|}{$<-6$mm} \\
    \cline{2-7}
    & $\pm0.3$mm & $\pm0.5$ mm & $\pm1.0$ mm & $\pm0.3$ mm & $\pm0.5$ mm & $\pm1.0$ mm \\
    \hline \hline
    1          & 62.02 & 75.12 & 85.58 & 10.87 & 26.99 & 48.54 \\
    2          & 56.55 & 72.52 & 85.14 & 05.44 & 17.67 & 46.21 \\
    3          & 56.03 & 71.28 & 84.15 & 03.11 & 19.81 & 43.69 \\
    4          & 47.04 & 64.86 & 81.84 & 00.78 & 05.83 & 34.56 \\\hline
    ConvLSTM   & 66.48 & 77.82 & 87.35 & 26.02 & 37.28 & 54.76 \\
    \hline
    \end{tabular}\label{tab:hypara_LSTM_accuracy}
    }\\
     \subfloat[GRU]{\begin{tabular}{|l||r|r|r|r|r|r|}
     \hline
     \multirow{3}{*}{Num. layers} & \multicolumn{6}{|c|}{Accuracy(\%) ($\uparrow$)} \\
    \cline{2-7}
    & \multicolumn{3}{|c|}{$<-4$mm} & \multicolumn{3}{c|}{$<-6$mm} \\
    \cline{2-7}
    & $\pm0.3$mm & $\pm0.5$ mm & $\pm1.0$ mm & $\pm0.3$ mm & $\pm0.5$ mm & $\pm1.0$ mm \\
    \hline \hline
    1          &  60.74 & 74.27 & 85.04 & 18.06 & 29.51 & 47.18 \\
    2          &  61.21 & 74.96 & 86.01 & 16.12 & 30.87 & 49.71 \\
    3          &  62.53 & 75.65 & 85.71 & 23.30 & 34.56 & 51.07 \\
    4          &  62.07 & 75.18 & 85.72 & 18.83 & 31.07 & 48.54 \\\hline
    ConvLSTM   &  66.48 & 77.82 & 87.35 & 26.02 & 37.28 & 54.76 \\
    \hline
    \end{tabular}
    }
    \caption{\label{tab:hypara_accuracy}Accuracy results by tuning the number of layers for (a) LSTM and (b) GRU.}
    \end{table}

\end{document}